\documentclass[final,3p,times,twocolumn,authoryear]{elsarticle}

\usepackage{amssymb}
\usepackage{graphicx}
\usepackage{subfig}
\usepackage{float}
\usepackage{overpic}
\usepackage{amsmath, amsthm, amssymb, graphicx}
\usepackage{booktabs, multirow}
\usepackage{caption}
\usepackage{bbm}
\usepackage{stfloats}
\usepackage{color}
\usepackage{hyperref}
\usepackage{CJKutf8}
\hypersetup{
    colorlinks=true,
    linkcolor=blue,
    filecolor=blue,      
    urlcolor=blue,
    citecolor=blue,
}
\usepackage{algorithm}
\usepackage{algorithmic}

\usepackage{xcolor}
\usepackage{soul}

\soulregister{\citep}7 
\soulregister{\citepp}7 
\soulregister{\citept}7 
\soulregister{\ref}7 
\soulregister{\pageref}7 
\usepackage{colortbl}  
\DeclareMathOperator{\vectorize}{vec}

\usepackage{lineno}

\long\def\comment#1{}

\definecolor{DarkGreen}{RGB}{81,158,62}
\newcommand\todocomment[1]{\textcolor{red}{#1}}
\newcommand{\jacomment}[1]{}

\makeatletter
\def\ps@pprintTitle{%
     \let\@oddhead\@empty
     \let\@evenhead\@empty
     \def\@oddfoot{\footnotesize\itshape\hfill}%
     \let\@evenfoot\@oddfoot}
\makeatother

\begin{document}
\begin{CJK}{UTF8}{gbsn}
\begin{frontmatter}




\title{Bayesian Fusion of Active Contour Models and ConvNet Priors for Standing Dead Tree Segmentation}


\author[label1]{Przemyslaw~Polewski}
\author[label2]{Jacquelyn~Shelton}
\author[label3,label4]{Wei~Yao\corref{cor1}}
\ead{wei.yao@ieee.org}
\cortext[cor1]{Corresponding author}
\author[label5]{Marco~Heurich}

\affiliation[label1]{organization={TomTom AI for Geospatial Research (TAIGR) group},
             addressline={TomTom North America},
             country={USA}}
\affiliation[label2]{organization={Department of Land Surveying and Geoinformatics},
             addressline={The Hong Kong Polytechnic University},
             city={Hung Hom, Kowloon},
             country={Hong Kong}}       
\affiliation[label3]{organization={Spatial Intelligence and Urban Computing},
             addressline={Institute of Urban Environment, Chinese Academy of Sciences},
             city={Xiamen},
             country={China}}
             
\affiliation[label4]{organization={State Key Lab for Ecological Security of Regions and Cities},
            addressline={Chinese Academy of Sciences},
             city={Xiamen},
             country={China}}
             
\affiliation[label5]{organization={Faculty of Environment and Natural Resources},
             addressline={University of Freiburg},
             country={Germany}}

\begin{abstract}
Instance segmentation is a core computer vision task with great practical significance and countless applications. 
%
Recent advances, driven by large-scale benchmark datasets, have yielded good general-purpose Convolutional Neural Network (CNN)-based methods.
However, many real-world problem domains grapple with imagery that have significantly different characteristics and structure. 
In particular, Natural Resource Monitoring (NRM) utilizes aerial remote sensing imagery with generally known scale and containing multiple overlapping instances of the same class, wherein the object contours are jagged and highly irregular, such as the crowns of individual trees in a dense forest.
This is in stark contrast with the more regular man-made objects found in classic benchmark datasets. 
%
%
In this work, we address this problem and propose a novel instance segmentation method geared towards NRM imagery. 
%
We formulate the problem as Bayesian maximum a posteriori inference which, in learning the individual object contours, incorporates shape, location, and position priors from state-of-the-art CNN architectures, driving a simultaneous level-set evolution of multiple object contours. We employ loose coupling between the CNNs that supply the priors and the active contour process, allowing a drop-in replacement of new network architectures. 
Moreover, we introduce a novel prior for contour shape, namely, a class of Deep Shape Models based on architectures from popular Generative Adversarial Networks (GANs). 
These Deep Shape Models are in essence a non-linear generalization of the classic Eigenshape formulation.
%
%
In numerical experiments, we tackle the challenging, real-world problem of segmenting individual dead tree crowns and delineating precise contours, which has significance for a multitude of ecological processes including carbon sequestration. 
We compare our method to two leading general-purpose instance segmentation methods -- Mask R-CNN and K-net -- on color infrared aerial imagery of dead tree crowns within the Bavarian Forest National Park.
Results show our approach to significantly outperform both methods in terms of reconstruction quality of tree crown contours.
Furthermore, use of our non-linear Eigenshape generalization, the GAN-based deep shape model prior, yields significant improvement of all results over the vanilla Eigenshape prior.
\end{abstract}

\begin{keyword}
Active contours \sep Instance segmentation \sep Aerial imagery \sep Bayesian inference


\end{keyword}

\end{frontmatter}


\section{Introduction}
Since the landmark introduction of the first scalable convolutional neural networks (CNN) into the computer vision mainstream~\citep{Krizhevsky2012}, CNN-based architectures have proliferated and achieved state-of-the-art (SotA) performance in many key vision tasks such as image classification~\citep{9609630}, object localization~\citep{conf/nips/RenHGS15} and semantic segmentation~\citep{RFB15a}. In addition, CNNs now form the backbone of generative adversarial network (GAN) architectures~\citep{DBLP:journals/corr/RadfordMC15}, which make it possible to learn low-dimensional models of image distributions that can nevertheless produce photorealistic image samples, sometimes indistinguishable from the originals. In case of semantic segmentation, the seminal work~\citep{Long2017} established a new paradigm of \emph{fully convolutional networks} (FCN), which did not contain fully connected layers and as such were capable of processing images of any dimension and outputting dense per-pixel maps instead of merely image-level labels. This enabled end-to-end training of the network on dense ground truth segmentation masks and led to the establishments of various forms of FCNs as today's SotA in semantic segmentation (see \cite{9356353} for a  survey). More recent surveys confirm that dead-tree and canopy-dieback mapping from UAV and aerial imagery remains an active and rapidly evolving research area, spanning applications from forest health assessment to biodiversity monitoring ~\citep{JUNTTILA2025419,chowdhury2026remote}.


The task of \emph{instance segmentation} can be seen as a combination of object detection and semantic segmentation. It forms the basis for many crucial computer vision applications dealing with autonomous driving, healthcare, medical imaging, augmented reality~\citep{GU2022104401} etc. The state-of-the-art approaches are CNN-based and may be roughly grouped into two categories: (i) \emph{top-down}, which first attempt to extract precise object bounding boxes and subsequently compute instance masks (e.g.~Mask R-CNN~\citep{8237584}), and (ii) \emph{bottom-up}, which start at the level of semantic segmentation and attempt to group/partition pixels into instances (e.g.~Deep Watershed Transform~\citep{8099788}). Arguably, most of the method development in the field has been driven by optimizing performance over large-scale benchmark datasets~\citep{coco2014,Geiger2012CVPR,ILSVRC15} dominated by terrestrial close-range photography (e.g.~pictures from social media and public image repositories). While this results in good performance on an 'average' image sampled from the unconstrained pool of web pictures, the general methods developed in this manner may be insufficient to segment imagery in some specialized domains.

In particular, natural resource monitoring (NRM) is an area of application that greatly benefited from rapid advancements in remote sensing imaging hardware.
However, the fine image details afforded by upgraded hardware have yet to be fully exploited by a new generation of algorithms that will cater to the characteristics of this imagery, which sets it apart from 'general-purpose' pictures. First, in NRM, the difficulty lies in separating many potentially overlapping instances of a single or few object classes (e.g. adjacent tree crowns in a satellite image), whereas standard benchmarks focus on the ability to separate hundreds of different classes. Interestingly, class-internal variation was named as one of the core difficulties of instance segmentation~\citep{8237584}. Second, remote sensing imagery is usually endowed with additional georeferencing or sensor trajectory information, which is not available in the general case and allows for some assumptions about object scale. Lastly, large-scale benchmarks are biased towards man-made objects with somewhat regular boundaries, oftentimes equipped with semantic constraints as to the location and orientation of the objects within the scene (e.g.~bottles must rest on a flat surface). Conversely, natural resources, e.g.~vegetation, are often characterized by an irregular and highly non-convex shape, and in the nadir view they usually do not possess a single valid orientation.

While some of the aforementioned problems could be tackled by training data augmentation techniques, perhaps the issue of intricate, irregular object boundaries poses an insurmountable challenge to convolutional networks. Indeed, deep CNNs are engineered for invariance to local image transformations, which allows them to learn hierarchical feature abstractions. However, this very trait intrinsically limits their precision of spatial localization~\citep{chen2014semantic,SANIMOHAMMED2022100024}. In the context of instance segmentation, the coarseness of CNN-derived feature maps can lead to degraded
performance for small and multi-scale object localization~\citep{DBLP:journals/corr/abs-1807-05511} and unsatisfactory object boundary prediction~\citep{E2EC_2022,isprs-archives-XLII-2-W16-127-2019}. Therefore, although CNNs have been established as the current state-of-the-art in instance segmentation, they alone may not be sufficient to achieve the desired level of accuracy and robustness.

Active contour models (ACMs) are a class of methods for evolving precise contours of objects within images according to an appropriate energy functional~\citep{KASS1987,Mumford1989OptimalAB}. In a sense, ACMs can be considered complementary to neural networks, because the former focus on localized information around an initial object contour, whereas the latter operate on the global image level build data abstractions that preserve invariance to local transformations. Therefore, it seems interesting to combine these two paradigms to produce an instance segmentation algorithm that can retain the best of both worlds. Indeed, the idea of a fused CNN/ACM based approach has been explored by a number of authors~\citep{Hatamizadeh2019,Marcos_2018_CVPR,POLEWSKI2021297,Le2017,Rupprecht2016,deepsnake2020}. To the best of our knowledge, no existing method satisfies the following conditions of automated instance segmentation: (i) handling of an arbitrary number of objects within the scene, (ii) fully automatic initialization of the contours, (iii) handling adjacent/overlapping objects and/or boundaries of same-class objects. The research presented in this paper aims to provide an integrated approach meeting the above criteria.

We propose an instance segmentation framework which performs statistical, multi-contour level-set evolution of shapes based on prior information integrated from state-of-the-art detection and semantic segmentation CNNs in a Bayesian fashion. The approach employs \emph{loose coupling} between the CNN and the ASM components, since they interact only through the prior information terms and thus can be trained independently. This is in contrast with existing fusion approaches, where the network architectures must be \emph{tightly coupled} with the evolving ASM of choice, and the training occurs jointly such that the ASM's error is back-propagated to the CNN. To enable loose coupling, we utilize a finite-dimensional representation of the evolving level-set function which inherently admits training, as opposed to the other approaches which rely on a variational representation requiring the coupled CNN to render it trainable. Specifically, we represent the level-set function by a vector of abstract shape coefficients, following the work of work of \cite{Leventon2010} and \cite{Tsai2003}. This permits the construction of an explicit shape prior on the shape coefficients from training object masks. The shape model can then be introduced into the Bayesian framework alongside the location and appearance information from the CNNs, turning the instance segmentation problem into maximum a posteriori inference. Finally, we propose an enhancement to the classic Eigenshape-based linear combination model of the evolving level set function. Instead of treating the shape vector elements as coefficients of the linear combination, we view them as latent variables which are fed into a decoder CNN based on the generator component of DCGAN~\citep{DBLP:journals/corr/RadfordMC15}, a popular generative adversarial network (GAN). This way, the generator network can be trained in a supervised fashion to decode shape coefficients back into the original image space, providing a richer and more expressive shape model compared to the usual sum-of-Eigenshapes formulation.

In summary, the main contribution of this work is an instance segmentation framework that
\begin{itemize}
    \item fuses prior information about appearance, location, and shape from state-of-the-art CNN models within a single Bayesian inference formulation,
    \item works with object locations (point), bounding boxes, or masks as initializations,
    \item employs \emph{loose coupling} between the evolving active contour part and CNNs that provide prior information, allowing drop-in replacements of networks without architectural changes,
    \item uses efficient contour initialization by projecting onto the learned shape model, leading to good segmentation results in the first attempt without random re-starts,
    \item leverages a novel shape model based on GAN architectures, paving the way for designing non-linear deep shape models, and 
    \item can be fully implemented and run on modern GPUs.
\end{itemize}
We validate our method on a challenging, real-world remote sensing problem: accurate segmentation of standing dead trees from very-high resolution aerial imagery. This task has ecological relevance due to the role of deadwood in carbon sequestration, forest nutrient cycles and biodiversity preservation~\citep{watsonExceptionalValueIntact2018, POLEWSKI2015252}. We compared our results against two baselines representing leading top-down (Mask R-CNN~\citep{8237584}) and bottom-up (K-net~\citep{zhang2021knet}) general-purpose instance segmentation methods and found that the active contour formulation can significantly improve the quality of reconstructed contours. Moreover, a comparison between the proposed DCGAN-based deep shape model and the classic Eigenshape linear model revealed that the former can further increase the contour precision. In this study, we focused on the single-class-multiple-instances case due to the application, however our methods extend naturally to the multi-class-multiple instances scenario, which is covered in the appendix.

This work builds upon and significantly extends our prior results~\citep{shelton2021hybrid,SANIMOHAMMED2022100024, Wong01112024}, where a sketch of the method was given. Extensions include introducing GAN-based deep shape models, a comprehensive experimental evaluation with additional baselines, and the generalization of the linear eigenshape model to arbitrary nonlinear kernels through the use of the decoder function. Beyond these methodological extensions, the present manuscript has also incorporated a review of instance segmentation approaches for tree crown and dead-tree delineation, and positioned our loosely-coupled Bayesian framework relative to recent foundation-model-based methods.

\section{Related work}
\comment{TODO clearly state the goal of each related work reference, the task they were trying to solve, how it's different}

\subsection{Fusion of active contours with CNNs}

There exist many generalizations of active contour methods to multiple evolving shapes. Early work includes the multi-phase Chan-Vese functional~\citep{VeseChan2002}, which uses separate binary phases to form an exponential encoding rule governing the assignment of pixels to regions. In~\cite{Samson1999}, each image region is assigned its own level set function, all of which are then evolved simultaneously. Alternatively,~\cite{1703607} proposed to recursively split binary regions, resulting in a hierarchical segmentation. More recently, research has focused on solving the multi-phase case using a single level-set function equipped with auxiliary functions to avoid gaps and overlap~\citep{Lucas2012} or constraints on region appearance such as the piecewise constant~\citep{Bae2009} or Gaussian mixture model~\citep{Dubrovina2015}.

\comment{rewrite this and re-analyze - copy paste from "Reformulating Level Sets as Deep Recurrent Neural Network Approach
to Semantic Segmentation"
\begin{itemize}
    \item Samson et al. (2000) associates a LS function
with each image region, and evolves these functions in a
coupled manner.
\item Lucas, B., Kazhdan, M., and Taylor, R. Multi-object spring
level sets (muscle). pp. 495–503, 2012. Lucas (2012) suggested
using a single LS function to perform the LS evolution for
multi-region segmentation. It requires managing multiple
auxiliary LS functions when evolving the contour, so that
no gaps/overlaps are created.
\item TODO include this - PAMI paper: \citep{Dubrovina2015} has developed an multi-region segmentation
with single LS function. ADD MORE DETAILS
\end{itemize}
}

Owing to its solid mathematical underpinnings, active contour methods and, in particular, their level-set realizations, can also be viewed from the perspective of their statistical formulation as a Bayesian inference problem~\citep{CremersIJCV}. This allows the explicit introduction of shape and appearance (image intensity) priors into the ACM functional. Early, general-purpose shape priors were effectively penalizing contour length, in the spirit of the minimal description length (MDL) rule (or broader, Occam's razor). However, in the pioneering work of~\cite{Leventon2010} and~\cite{Tsai2003}, a learnable, data-driven shape prior was developed based on Principal Component Analysis (PCA; \cite{Hotelling1933}) of level-set functions encoding the training shapes. Thus, the \emph{variational} problem of finding the optimal level-set function $\phi$ was reduced to a low-dimensional \emph{optimization} problem of finding the multiplicative eigenmode coefficients which optimally express the image's segmentation. Initially, Gaussian~\citep{Leventon2010} or uniform~\citep{Tsai2003} distribution on the coefficients was assumed, which turned out to be too restrictive, leading to the extension by Cremers and Rousson~\citep{Cremers2007}, utilizing a non-parametric kernel density estimator model.


\jacomment{*** we'll make this consistent with everything we present from here on -- refer back to it with e.g. fig/description of U-net, Mask R-CNN, indicating which part of the figure / arrows in the figures represent.}
\begin{figure*}[t!]
\centering
		\includegraphics[width=0.7\textwidth]{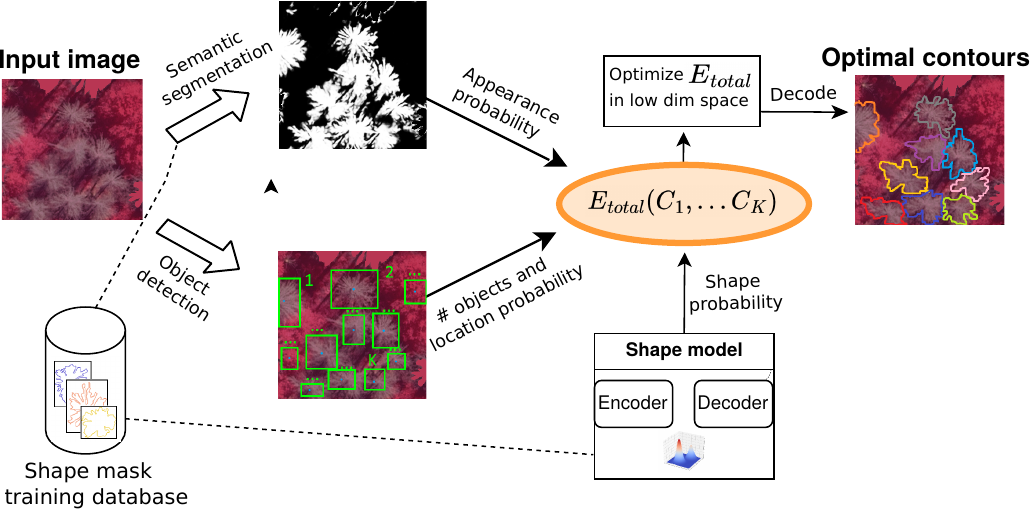}
	\caption{Illustration of the full pipeline of our proposed instance segmentation approach.
Our method embeds information from two convolutional neural networks, namely \textit{semantic segmentation} and \textit{object detection} networks.
This is performed in a multi-contour simultaneous optimization scheme over abstract, low-dimensional shape coefficients, to obtain high-quality object segmentations. \jacomment{Fix fig, obvs, and label different networks employed to get to that next representation, **see nips workshop paper}}
	\label{fig:strategyOverview}
\end{figure*}

\comment{
Part 2 - introducing prior information into active contours?
\todocomment{do we want this here too??}
\begin{itemize}
    \item work of Leventon et al., Tsai et al. - eigenshapes
    \item Cremers and Rousson -> KDE on eigenshape coefficients
    \item \todocomment{copy paste from "Reformulating Level Sets as Deep Recurrent Neural Network Approach
to Semantic Segmentation": shape prior
(Le \& Savvides, 2016) have been also considered.}
\end{itemize}
}

We now turn to prior work directly concerned with combining CNN architectures with active contour models. Several authors propose to generalize a scalar parameter of the original 'unsupervised' Chan-Vese functional~\citep{ChanVese2001} to a 2D raster, the value of which is then calculated by a CNN based on the input image being segmented. The CNN is trained in conjunction with the active contour model by back-propagating a suitable loss on the final segmentation result through the network's layers. This approach is followed by~\cite{Marcos_2018_CVPR} with respect to the geometric prior and contour length penalty terms. Also,~\cite{Hatamizadeh2019} utilize a similar approach to learn the initial level set function $\phi_0$ as well as maps of the originally scalar regularization parameters $\lambda_1,\lambda_2$ which control the balance between foreground and background during the contour evolution. In a different approach~\citep{Le2017}, the active contour method is recast in a recurrent neural network-based (RNN) framework. Notably, the evolution of the level-set function is governed by a gated recurrent unit (GRU), where the additional matrix parameters are trained by back-propagation through time of the per-pixel cross-entropy loss on the segmented image versus ground truth. The aforementioned approaches all share the trait of evolving the signed-distance function $\phi$ explicitly, as opposed to our method which utilizes a low-level representation of $\phi$ in the form of abstract shape coefficients, the distribution of which can be modeled directly. Also, these methods do not make any attempt at penalizing the overlap or modeling the boundary of adjacent shapes, which may be crucial for difficult scenes containing many overlapping shape instances from the same class (as in the case of our application).

Another group of approaches for fusing active contours with CNNs is built around the idea of explicitly representing the object contour through its vertices, and iteratively estimating a \emph{displacement field} by a CNN, resulting in an offset vector for every contour vertex which is applied to it as part of one iteration of the evolution. Early examples include~\cite{Rupprecht2016}, where the CNN predicts offset vectors to the closest contour point of a training object based on fixed-size patches around the evolving contour. The initial contour must be provided by an 'oracle'. Another example of this class of approaches is~\cite{Gur2020End}, where an encoder-decoder CNN architecture is used to predict the displacement field of the evolving polygon's vertices. A neural mesh renderer~\citep{Kato_2018_CVPR} is used to turn the discrete polygonal vector representation into a raster, which can then be used to back-propagate the per-pixel loss over ground truth segmentation masks. This method also requires an 'oracle'-provided initialization of the contour.
In the DeepSnakes approach~\citep{deepsnake2020}, the contour is initialized to $N$ vertices sampled from the bounding octagon, and the CNN extracts offsets to the object boundary based on convolutional feature maps and the circular convolution operator. Recently, a learnable contour initialization framework coupled with global contour deformation was proposed to address some of the training and convergence problems of DeepSnakes~\citep{E2EC_2022}. The presented class of methods that use the explicit vertex representation differs from our approach, and in a broader sense from the previously discussed $\phi$-evolution methods, in that it mostly considers local information around the contour. Therefore, poor initializations with contours away from the true object boundary could lead to unsatisfactory results. Also, these methods might be less applicable to remote sensing imagery, where the quality and saliency of object boundaries is not always as uniform as with high-resolution terrestrial photography, which forms the basis of many modern benchmark datasets for image segmentation~\citep{coco2014,Geiger2012CVPR,ILSVRC15}.
For a comprehensive treatment of CNN-based segmentation methods, including fusion with active contours, see the recent review by~\cite{9356353}.

Finally, our method builds on some ideas from the work of~\cite{Bakir2003}, where it was proposed to 'decode' representations in kernel space using machine learning methods as opposed to the direct analytical or optimization-based route. Notably, they utilized kernel ridge regression to learn this mapping, whereas we apply the generator network from the DCGAN~\citep{DBLP:journals/corr/RadfordMC15} architecture.

\subsection{Recent advances in tree crown and dead-tree instance segmentation.}

Beyond the CNN-ACM fusion methods discussed above, a substantial body of recent work has targeted individual tree crown delineation directly with CNN- or transformer-based instance segmentation architectures. These include YOLO-based detectors adapted for tree crowns~\citep{SUN2026103748}, star-convex polygon representations built on U-Net ~\citep{TONG2025114618}, multi-task networks that jointly perform crown delineation, species classification, and vitality assessment~\citep{FAN2026261}, and Mask R-CNN variants augmented with multi-source UAV multispectral and canopy-height data through adaptive fusion modules~\citep{SUN2026103748}. Of particular relevance to our multi-contour interaction formulation, ~\cite{Do2026} propose TreeCoG, a bottom-up approach that deliberately over-segments tree crowns via contour detection and subsequently merges the resulting contour candidates using a graph convolutional network built on pairwise contour relationships — a strategy conceptually related to, but fundamentally different from, our probabilistic graphical model over shape, location, and overlap potentials, which merges evidence in the energy domain rather than through explicit graph-based contour clustering.

A parallel line of work has adapted large-scale vision foundation models, most notably the Segment Anything Model (SAM), to the forestry domain. These approaches fine-tune SAM in a parameter-efficient manner for tree-species segmentation ~\citep{Xu2026}, couple SAM with YOLO-based detection for joint crown delineation and species classification~\citep{QUE2026111162}, or incorporate elevation cues to guide SAM-based crown segmentation from drone imagery~\citep{NEURIPS2025_0cb66bdf}. At a broader scale, multi-modal, multi-task foundation models have been proposed for forest monitoring pipelines as a whole ~\citep{bountos2025fomo}, while flow-guided segmentation has been explored as a means of delineating tree crowns without requiring instance-level annotations~\citep{11520829}. Compared with our framework, these foundation-model-based approaches typically rely on tight or prompt-based coupling between the segmentation backbone and the downstream task, whereas our Bayesian formulation is explicitly designed to accommodate any drop-in CNN prior — including, in principle, foundation-model-derived priors — without altering the active-contour evolution itself.

Finally, complementary to per-image contour delineation, several recent studies address the temporal dimension of tree mortality using satellite image time series, e.g. combining Sentinel-2 time series with Random Forest and U-Net classifiers to map bark-beetle-induced dieback ~\citep{10.1145/3605098.3635908}, or applying change-point detection to PlanetScope time series to estimate the timing of individual tree deaths~\citep{NOWAK2026261}. While such time-series approaches operate at coarser spatial resolution than the aerial imagery considered here, they highlight a growing interest in characterizing not only the spatial extent but also the temporal dynamics of forest disturbance

\section{Proposed framework}

We begin by defining the notation and formalizing the problem setting. Consider the image plane $\Omega \subset \mathbb{R}^2$ and a $d$-dimensional image $I: \Omega \mapsto \mathbb{R}^d$. We assume that an unknown number $N_o$ of \emph{objects} from a known set of classes $\mathcal{C}, |\mathcal{C}| \geq 1$ is present within the image $I$.
For \emph{instance segmentation}, the goal is to retrieve a set of pairwise-disjoint masks 
\begin{equation}
\mathcal{M}=\bigcup_{i=1}^{N_o}M_i, \forall_{i \neq j} M_i \cap M_j = \emptyset
\end{equation}
which represent the assignment of regions within the image to the objects.

An overview of our strategy for recovering the object masks from a new test image $I^t$ is depicted in Fig.~\ref{fig:strategyOverview}. First, a semantic segmentation network and an object detector, both pre-trained on a set of training shapes, are applied to $I^t$. Next, the semantic segmentation map and initial object locations are used in conjunction with an object shape model, also learned from training masks, to construct a smooth energy functional on the space of evolving shape contours. The shape model includes an encoder and a decoder, where the task of the former is to \emph{encode} an object mask into a compact, low-dimensional representation, while the \emph{decoder} performs the inverse operation. The total energy, encompassing the object appearance, shape, and location, is then minimized with respect to the encoded low-level contour representations with gradient-based methods. After decoding, the total energy yields the target masks of the objects of interest present within the image $I^t$, thereby segmenting each \emph{instance} of the objects.

In order to construct our aggregate framework for estimating $\mathcal{M}$, we will first present and review the classic \emph{single-contour} statistical formulation of active contour methods, and subsequently build on the introduced concepts to show our generalization to multiple interacting contours.

\subsection{Single-contour statistical level-sets}

\begin{figure}[h!]
\centering
   \subfloat[evolving contour\label{1a}]{%
       \includegraphics[width=0.38\linewidth]{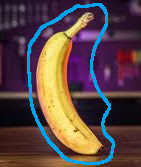}}
    \hfill
  \subfloat[signed distance to contour\label{1b}]{%
        \includegraphics[width=0.61\linewidth]{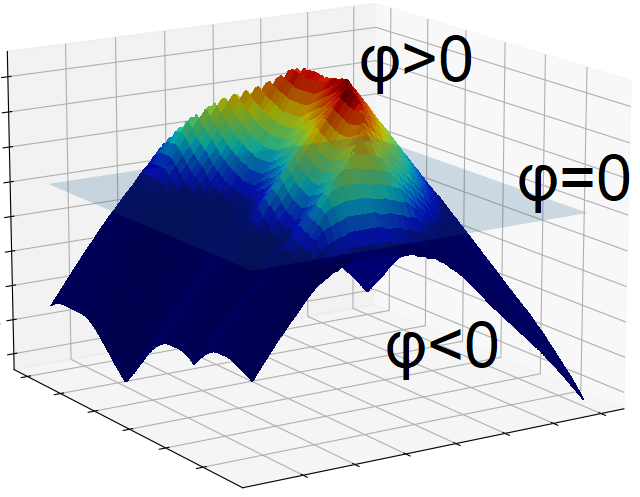}}
	\caption{Left: an evolving contour (in blue) partitions the image plane into the foreground and background regions. Right: Corresponding level-set function representation of evolving shape. The contour corresponds to the function's zero level set, while positive and negative function values represent, respectively, image elements inside and outside the contour.}
	\label{fig:evolvingLevelSetExample}
\end{figure}

The classic statistical formulation of active contour segmentation assumes that an evolving front (contour) $C$ partitions the image plane into two disjoint parts $\Omega_F, \Omega_B \subseteq \Omega$, respectively the \emph{foreground} and \emph{background} regions. The foreground region delineates the segmented object as visible within the image. 
The posterior probability of observing a partition $\Omega_F,\Omega_B$ given image $I$ can be described by the Bayes rule as proportional to the likelihood of observing $I$ given the partition and the \emph{prior probability} of the partition~\citep{CremersIJCV}:
\begin{equation}
\mathcal{P}(\Omega_F,\Omega_B|I) \propto
\mathcal{P}(I|\Omega_F,\Omega_B)\mathcal{P}(\Omega_F,\Omega_B)
\label{eq:bayesianRegions}
\end{equation}
The contour evolution is driven by the maximum a posteriori inference on criterion~\ref{eq:bayesianRegions}. Alternatively, this criterion can be expressed equivalently in terms of \emph{energy} of the contour $E(C)=-\log \mathcal{P}(\Omega_F,\Omega_B|I)$, since $C$ uniquely identifies the partition $\Omega_F,\Omega_B$. Assuming that region labellings are uncorrelated, i.e.
$\mathcal{P}(I|\Omega_F,\Omega_B)=\mathcal{P}(I|\Omega_F)\mathcal{P}(I|\Omega_B)$,
and also that image values inside a region are independent and
identically distributed realizations of one underlying random process~\citep{CremersIJCV}, the image energy term $E_{img}(C)=-\log \mathcal{P}(I|\Omega_F,\Omega_B)$ becomes:
\begin{equation}
\small
\label{eq:eimg}
E_{img}(C) = -\int_{\Omega_F} log f_F(I(x))dx-\int_{\Omega_B}
log f_B(I(x))dx
\end{equation}
\normalsize
In the above, the terms $f_F, f_B$ indicate the probability density functions of image intensities within, respectively, the foreground and background regions. The symbol $\mathbbm{1}_F$ denotes the indicator function for membership in the foreground region. It is often convenient to represent the evolving contour as the zero \emph{level-set} of an appropriate function $\phi$, such that $\phi > 0$ for elements inside the evolving shape,  $\phi < 0$ for elements outside the contour (i.e.~belonging to the background), and $\phi = 0$ on the contour itself~\citep{OSHER198812} (see Fig.~\ref{fig:evolvingLevelSetExample} for an example). In this case, the indicator $\mathbbm{1}_F$ can be expressed by the Heaviside function: $\mathbbm{1}_F[x] \equiv H(\phi[x])$. Using basic algebra, Eq.~\ref{eq:eimg} can be reformulated as:
\begin{equation}
\begin{split}
\label{eq:eimg_phi}
E_{img}(\phi) &= -\int_{\Omega} H(\phi[x])\log \frac{f_F(I(x))}{f_B(I(x))} dx + Z
\end{split}
\end{equation}
The value $Z$ is a constant with respect to the evolving contour $\phi$ and as such does not influence the location of the optimum. Furthermore, it is possible to reformulate Eq.~\ref{eq:eimg_phi} in terms of only the discriminative probability $\mathcal{P}(x \in \Omega_F|I(x))$ as opposed to the generative per-region probabilities $f_F, f_B$ by applying the Bayes theorem to both the numerator and denominator of the ratio $f_F(I(x))/f_B(I(x))$ and assuming an equal class occurrence prior $\mathcal{P}(x \in \Omega_F) = \mathcal{P}(x \in \Omega_B)$~\citep{isprs-annals-II-3-W4-181-2015}. We obtain the following form of the term $E_{img}$:
\begin{equation}
\small
\begin{split}
\label{eq:eimg_discrim}
E_{img}(\phi) &= -\int_{\Omega} H(\phi[x])\log \frac{\mathcal{P}(x \in \Omega_F|I(x))}{1-\mathcal{P}(x \in \Omega_F|I(x))} dx + Z
\end{split}
\normalsize
\end{equation}
It should be noted that minimizing $E_{img}$ is equivalent to minimizing average cross-entropy between the assignment of image elements to the foreground/background and their posterior class probabilities over the entire image $I$. To see this, define $E'_{img} = E_{img} - \int_{\Omega}\mathcal{P}(x \in \Omega_B|I(x))$, which differs from $E_{img}$ only by a constant. This results in the familiar binary cross-entropy:

\begin{equation}
\begin{split}
\label{eq:eimg_crossEntr}
E'_{img}(\phi) &= -\int_{\Omega} H(\phi[x])\log \mathcal{P}(x \in \Omega_F|I(x)) +\\
&\qquad (1-H(\phi[x]))\log [1-\mathcal{P}(x \in \Omega_F|I(x))]dx
\end{split}
\end{equation}

In the classic level-set approach, the shape prior $\mathcal{P}(\Omega_F,\Omega_B) \equiv \mathcal{P}(\phi)$ is usually formulated in the spirit of the minimal description length rule~\citep{10.1093/comjnl/11.2.185,RISSANEN1978465}, penalizing the length of the interface (contour):
\begin{equation}
    \mathcal{P}(\phi)=\exp(-\nu |\nabla H(\phi)|)
\end{equation}
In the next section, we will review the developments leading to the establishment of a method for training shape priors from annotated images, which we build upon in our framework.

\subsection{Representing the level-set function compactly}

Optimizing the level-set segmentation criterion~\ref{eq:bayesianRegions} is a \emph{variational}, infinite-dimensional problem - we seek the optimal function $\phi$ minimizing the appropriate statistical functional. As such, it is difficult to construct statistical models of $\phi$ directly. To circumvent this problem, certain \emph{indirect}, low-dimensional representations of $\phi$ have been developed which admit modeling based on training samples. In particular, the pioneering work of Leventon et al.~\cite{Leventon2010} and~\cite{Tsai2003} established a way of representing the level-set function $\phi$ as a linear combination of \emph{eigenshapes}, i.e.~eigenvectors of a shape variability matrix which was constructed from binary masks of labeled training shapes. Specifically, given a set $\mathcal{M}^T=\{M^T_1,...,M^T_{N_T}\}$ of $W \times H$-sized binary training masks representing the shapes of objects from the target class, the first step is to construct the level-set representations $\{\phi^T_1,...,\phi^T_{N_T}\}$ of the training shapes by extracting the signed distance from the contour. The \emph{shape variability matrix} is formed by concatenating the vectorized, de-meaned individual signed distance matrices:
\begin{equation}
\label{eq:shapeVarMatrix}
S=\begin{pmatrix}
\vectorize(\phi^T_1- \bar{\phi}) \,\,\, \vectorize(\phi^T_2- \bar{\phi})  \,\,\,\cdots \,\,\, \vectorize(\phi^T_{N_T}-
\bar{\phi}) \\
\end{pmatrix}
\end{equation}
Singular value decomposition of $S$ is then performed, yielding the left-singular vectors (equivalently, eigenvectors of $SS^T$) $\Psi_1,...,\Psi_c$ corresponding to the top $c$ singular values. This gives rise to the following parametric, finite-dimensional representation of $\phi$ in terms of a coefficient vector $\vec{\alpha}=[\alpha_1, ...,\alpha_c]^T$ which encodes each eigenshape vector's magnitude of influence within the linear combination:
\begin{equation}
\label{eq:phi_eigenshape}
    \phi(\vec{\alpha})=\bar{\phi} + \sum_{i=1}^c \alpha_i\Psi_i
\end{equation}
The above representation yields the level-set function $\phi$ in \emph{standard} position and orientation,  i.e.~the mask centroids are located at the origin of the coordinate system. In order to enable alignment of the evolving shape's position, size and rotation with the evidence contained in the segmented image in the course of the contour evolution, an affine transformation $\mathcal{T_\rho}$ is applied to the coordinates $x$ of the target image element $\omega \in \Omega$ in the original image space:
\begin{equation}
\label{eq:phi_repr_xform}
    \phi(x|\vec{\alpha},\rho)=\bar{\phi}[\mathcal{T_\rho}x] + \sum_{i=1}^c \alpha_i\Psi_i[\mathcal{T_\rho}x]
\end{equation}

\begin{figure}[h!]
\centering
   \subfloat[uniform model\label{1a}]{%
       \includegraphics[width=0.49\linewidth]{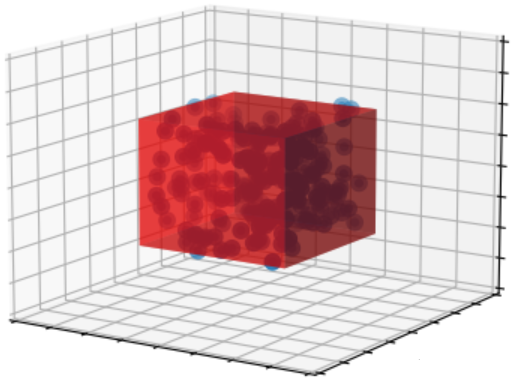}}
    \hfill
  \subfloat[KDE model\label{1b}]{%
        \includegraphics[width=0.49\linewidth]{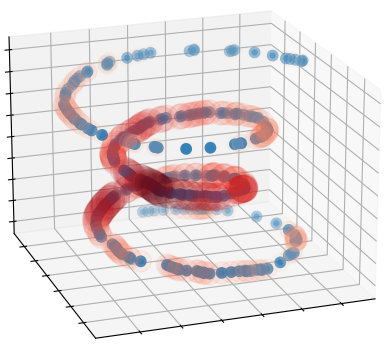}}
        
	\caption{Visualization of the concept of a uniform vs. KDE probability model for shape coefficients. Left: uniform model assigns equal probability to all combinations of shape coefficients (blue points) in the hypercube spanned by the maximum extents of training examples. Right: KDE model assigns high probability only to regions around training samples (blue points), with arbitrary topology. The intensity of the red- colored transparent spheres around the blue points is proportional to the probability density of the shape coefficients in that region (dark red indicates highest probability.}
	\label{fig:typesOfShapeProbModels}
\end{figure}

By representing the level-set function $\phi$ in terms of a low-dimensional coefficient vector $\vec{\alpha}$, the \emph{variational} problem of finding the optimal $\phi$ has now been transformed into an \emph{optimization} problem of maximizing the a posteriori probability over $\mathbb{R}^c$. Moreover, since the representation can be made invariant under the desired affine transformations (see Eq.~\ref{eq:phi_repr_xform}), the $\vec{\alpha}$ coefficients encode the 'absolute' shape variation range of the target object class. This allows constructing \emph{models} of the shape coefficients based on the representations $\{\vec{\alpha}^T_1,...,\vec{\alpha}^T_{N_T}\}$ of available training masks $\mathcal{M}^T$. The Gaussian~\citep{Leventon2010} and uniform~\citep{Tsai2003} distributions on the coefficients were among the first proposed models. Later, Cremers and Rousson~\citep{Cremers2007} provided a more general, non-parametric model on the shape coefficients by constructing a kernel density estimator on the similarity between a sample $\vec{\alpha}$ vector of coefficients and the training coefficients $\vec{\alpha}^T_i$ as measured by an appropriate kernel $\mathcal{K}_\alpha$:
\begin{equation}
    \label{eq:shape_prob_alpha_coef}
    \mathcal{P}_{shp}(\vec{\alpha}) \propto \sum_{i=1}^{N_T}\mathcal{K}_\alpha(\frac{||\vec{\alpha} - \vec{\alpha}^T_i||}{\sigma})
\end{equation}
The choice of kernel bandwidth $\sigma$ determines the size of the neighborhood around training points where the shape coefficient probability remains high. See Fig.~\ref{fig:typesOfShapeProbModels} for a visualization of these ideas.

\subsection{Generalizing the eigenshape representation}\label{sec:generalizing_eigenshape_rep}

In this section, we extend the classic eigenshape based representation of the level-set function in a way which can (i) overcome a certain theoretical difficulty and (ii) offer the possibility of applying more expressive shape models and task-specific representations. Regarding (i), it is known that the additive model from Eq.~\ref{eq:phi_eigenshape} is theoretically unfounded, because the sum of signed distance functions does not, in general, constitute a signed distance function itself~\citep{Tsai2003}. To that end, we propose to treat the vector $\vec{\alpha}$ as a sequence of abstract shape coefficients, without any direct relationship to any specific component of a pre-trained shape model. Indeed, $\vec{\alpha}$ is now simply a vector parameter for the \emph{decoder function} $f_d(\vec{\alpha}) \approx \phi_{\vec{\alpha}}$, which approximates the level-set function $\phi$ corresponding to the shape coefficient combination encoded by $\vec{\alpha}$. Note that in this case, $\phi$ is not constrained to be the contour signed-distance function - it can be any function such that $H(\phi)$ constitutes a good approximation to the evolving contour. The decoder function provides an interface for incorporating rich shape models encapsulating arbitrary non-linearities. In Section~\ref{sec:deepShapeModels}, we show how a class of convolutional neural networks can be efficiently used in this role. We place two constraints on the decoder function:
\begin{enumerate}
    \item \emph{trainability}: $f_d(\cdot)\equiv f_d(\cdot|\theta)$, where $\theta$ is the vector of trainable parameters
    \item \emph{smoothness}: both $\frac{\partial f_d}{\partial \vec{\alpha}}$ and $\frac{\partial f_d}{\partial\theta}$ should exist
\end{enumerate}


The smoothness requirement allows optimization with gradient-based methods, which makes our framework implementable on modern GPU-based computation frameworks. On the other hand, trainability ensures that the shape model may be learned from labeled training images (i.e.~the training masks $\mathcal{M}^T$). However, after abandoning the signed-distance/eigenshape representation, there remains the open question of how to obtain the shape coefficients $\{\vec{\alpha}^T_1,...,\vec{\alpha}^T_{N_T}\}$ of the training masks. To this end, we propose to define the topology of the coefficient space $A \subset \mathbb{R}^c$ based on the similarity relationships present in the training data, as measured by a kernel function $\mathcal{K}_T$. Specifically, $\mathcal{K}_T(\cdot,\cdot)$ acts on two binary training masks $M_1, M_2 \subset \mathbb{R}^{W \times H}$ in 'standard' position (normalized with respect to the desired affine transformations) and produces a non-negative real number quantifying the similarity between $M_1$ and $M_2$. Thus, the matrix $K_T=[\mathcal{K}(M^T_i,M^T_j)]_{i,j=1..N_T}$ is constructed and diagonalized as per the kernel principal component procedure proposed by~\cite{scholkopf1997kernel}. Provided that the kernel $\mathcal{K}_T(x,y)$ satisfies Mercer's condition~\citep{MR0188745}, it is the inner product of an implicit, higher-dimensional space $F$ given by the (in general unknown) transform $\Phi(\cdot)$: $\mathcal{K}_T(x,y) = \Phi(x) \cdot \Phi(y)$. 
\jacomment{use notation from kernel book / kernel papers for kernel mapping func}
Therefore, diagonalizing $K_T$ amounts to the eigendecomposition of the lifted-space covariance matrix $\mathcal{C}=\frac{1}{N_T}\sum_{i=1}^{N_T} \Phi(M^T_i)\Phi(M^T_i)^T$. The eigenvectors can be represented as linear combinations of the transformations $\Phi(M^T_i)$ of the original training data:
\jacomment{something about why this is guaranteed, so long as our kernel function meets the KKT conditions or whatever (CITE), then it's coefficients admit the form bla according to the Representer theorem (CITE).  Get notation/wording from e.g. Lampy, Arthur, Matthew papers.}
\begin{equation}
\label{eq:kernel_eigv}
    V = \sum_{i=1}^{N_T} \beta_i \Phi(M^T_i)
\end{equation}
Following~\cite{scholkopf1997kernel}, we consider the constrained eigenvalue problem of the form\footnote{In practice, the kernel matrix is first centered in the space $F$, yielding $\tilde{K}=K_T-1_{N_T}K_T-K_T 1_{N_T}-1_{N_T}K_T 1_{N_T}$, where $(1_x)_{i,j} = 1/x$}:
\begin{equation}
\label{eq:kernel_eigv_problem}
\begin{split}
&N_T \lambda \beta = K_T \beta \\
&\forall_{k: \lambda^{(k)} \neq 0} 1 = \beta^{(k)} K_T \beta^{(k)}
\end{split}
\end{equation}
In the above, the condition $\beta^{(k)} K_T \beta^{(k)}$ for each non-zero eigenvalue separately implies that all the eigenvectors in $F$ are normalized, since by Eq.~\ref{eq:kernel_eigv}, $V\cdot V=\beta K_T \beta=1$. To extract the representation of a point $x \subset \mathbb{R}^{W \times H}$ in terms of the principal components determined by $K_T$, we compute the projection of $\Phi(x)$ onto the first $c$ eigenvectors obtained from Eq.~\ref{eq:kernel_eigv_problem}, sorted by the magnitude of the corresponding eigenvalues $\lambda$. The projection onto a single eigenvector $V^k$ can be written as:
\begin{equation}
\label{eq:projectingShapeCoefs}
    (V^k \cdot \Phi(x)) = \sum_{i=1}^{N_T}\beta_i^k \mathcal{K_T}(M^T_i, x)
\end{equation}
We can now define the shape coefficients $\alpha^T_i$ associated with the training masks from $\mathcal{M}^T$:
\begin{equation}
\label{eq:shapeCoefTrain}
    \alpha^T_i = \left[ V^1 \cdot \Phi(M^T_i), \dots, V^c \cdot \Phi(M^T_i) \right]
    \quad \forall_{i=1, \dots, N_{T}}  
\end{equation}
The set of pairs $(M^T_i, \alpha^T_i)$ constitutes the training set for the decoder function $f_d(\cdot|\theta)$. To ensure the scalability of the proposed approach, we would like to point out that incremental algorithms for performing kernel PCA are available (e.g.~\citep{https://doi.org/10.48550/arxiv.1802.00043}), enabling support for possibly millions of training shapes and circumventing the cubic computational complexity of 'one-shot' kernel PCA.

Finally, a key difference between the kernel PCA approach and the original eigenshape-based representation should be emphasized. Note that in the former case, the actual eigenvectors $V^k$ in the lifted feature space $F$ are never explicitly constructed. Indeed, they might not be obtainable at all due to the fact that $F$ may be infinite-dimensional. Therefore, it is in general impossible to construct an additive representation with explicit references to eigenvectors of shape variation, resembling Eq.~\ref{eq:phi_eigenshape}. Instead, the kernel PCA procedure serves as more of a black-box extractor ('encoder') which provides a low-level representation of the space of training shape masks. However, this is not a problem, because the decoder function \emph{learns} the inverse mapping from the low-level representation back to the shape mask space, in the spirit of~\cite{Bakir2003}, without the need to ever manipulate eigenvectors explicitly.

\subsection{Simultaneously evolving multiple shapes}\label{sec:evolvingMultipleShapes}

\begin{figure}[h!]
\centering
	\subfloat[original image]{%
       \includegraphics[width=0.32\linewidth]{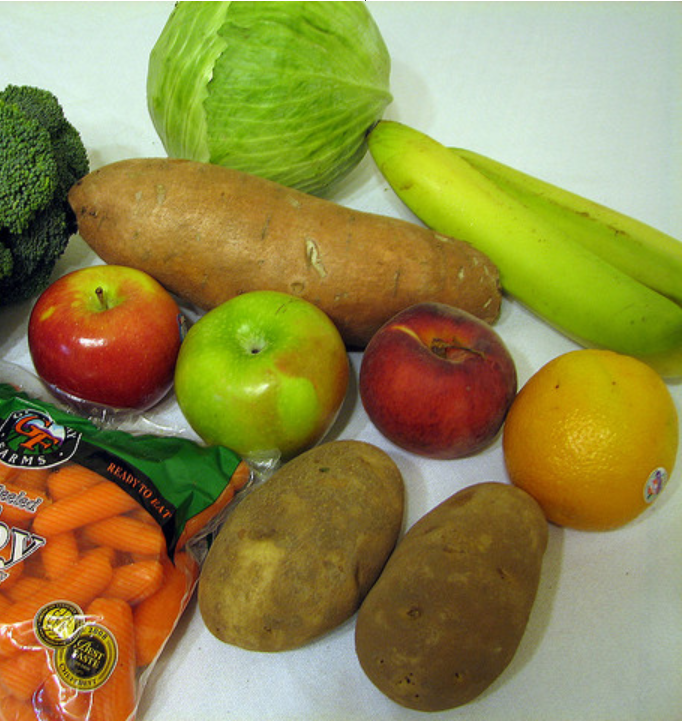}}
    \hfill
  \subfloat[3-class semantic segmentation]{%
       \includegraphics[width=0.32\linewidth]{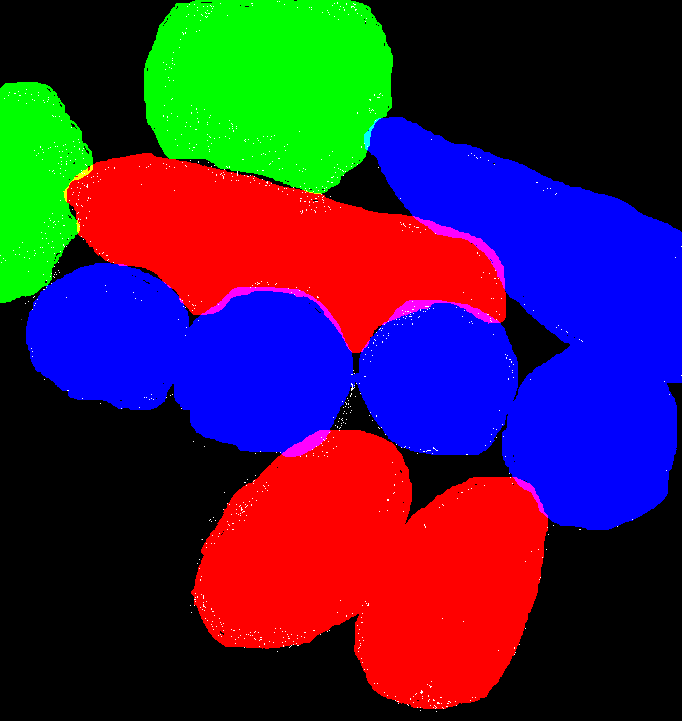}}
    \hfill
    \subfloat[bounding boxes]{%
       \includegraphics[width=0.30\linewidth]{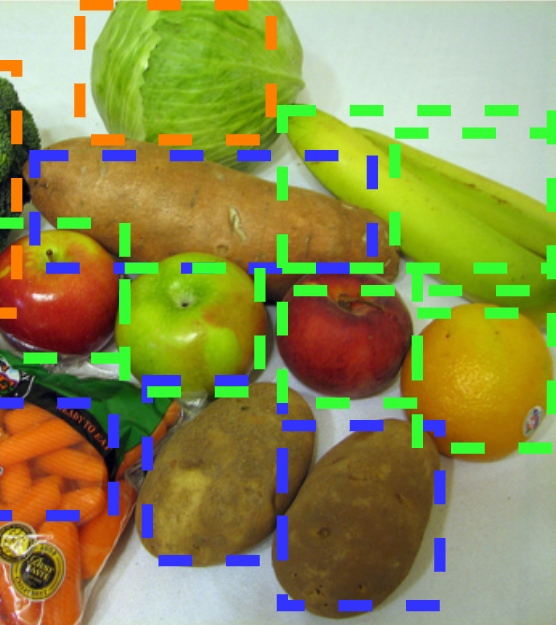}}
	\caption{Sample input for a 3-class segmentation problem: 
        green vegetables (shown in \textcolor{green}{green}), potatoes (\textcolor{red}{red}), and fruit (\textcolor{blue}{blue}).
       Shown: (a) original RGB image containing examples of all classes of objects, (b) 3-class (+background) semantic segmentation, (c) initial detected object bounding boxes. (Images taken from the COCO dataset.)}
	\label{fig:veggiesSingleCol}
\end{figure}

Having defined some key ingredients of our active contour approach, notably the kernel PCA based abstract shape representation, the decoder function $f_d$ mapping the representation back into the image space, and the associated non-parametric probability model, we now proceed to introduce the remaining building blocks and concepts. First, building on the recent advancements in the application of CNNs for semantic image segmentation as well as object detection/localization, we assume the following as input to our framework (see Fig.~\ref{fig:veggiesSingleCol}):

\begin{itemize}
    \item a dense \emph{semantic probability model} $\mathcal{P}_{\text{sem}}(c|\omega,I)$, quantifying the posterior class probability distribution over the possible classes $\mathcal{C}$ for each image element $\omega$ given the image evidence $I$ 
    \item a set of $\tilde{N_o}$ approximate initial object center positions $\tilde{\omega}^{ctr}_{i}$, along with their detected class $c_i$ and approximate bounding box $\tilde{M}_i$
\end{itemize}

The semantic model can be thought of as the output of a state-of-the-art fully convolutional network for semantic segmentation, although in principle any dense posterior probability map could be applied. The set of approximate initial positions can be obtained e.g.~from an object detection network. It is more beneficial for the initial number of objects $\tilde{N_o}$ to be overestimated than underestimated, because under our framework an oversegmentation error could still be recovered from by the redundant shapes evolving to an empty contour, whereas a missing initial shape will not be accounted for. The bounding boxes $\tilde{M}_i$ need not be precise, and may just be derived from average bounds of objects belonging to class $c_i$. 

We make use of the following abstractions which are the final two customizable components of our framework and must be specified before instantiation: 

\begin{itemize}
    \item an \emph{object location model} $\mathcal{P}_{loc}(\omega|\tilde{\omega}^{ctr},c)$, which describes the probability of an object's center position being located at image element $\omega$, given the object's class and initial position $\tilde{\omega}^{ctr}$
    \item an \emph{object orientation model} $\mathcal{P}_{rot}(\kappa|c)$, which encodes the per-class probability of observing an object instance having rotation $\kappa$
    \item a pairwise \emph{interaction model} $\mathcal{P}_{int}(|M_i \cap M_j|;c_i,c_j)$ which restricts the potential overlap between two shapes $M_i,M_j$ of possibly different classes
\end{itemize}

\begin{figure*}[h!]
\centering
		\includegraphics[width=0.7\textwidth]{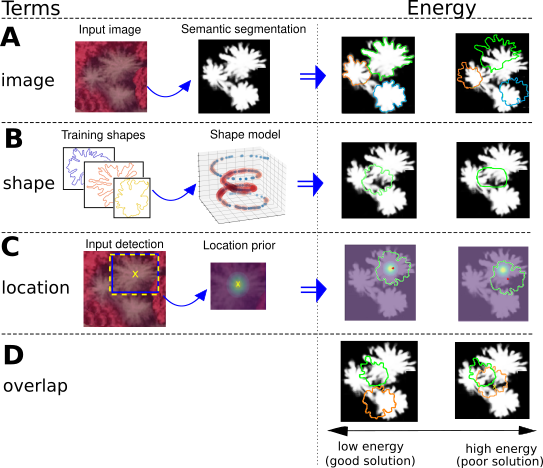}



	\caption{\jacomment{jackie primp}Visual comparison of shape/location parameter configurations which lead to low vs.~high values of partial energy terms contributing to total energy from Eqs.~\ref{eq:multiContourEnergy},\ref{eq:E_prior_terms}. Evolving contours shown as green, orange, blue polygons over semantic segmentation map. Consider the segmentation problem of delineating individual dead tree crowns (see Section~\ref{sec:problemSetting} for details).
    \textbf{A}: input (original) RGB image transformed into per-pixel probability map of belonging to the dead tree class. right: image term is low when high-probability class pixels lie mostly within the contours and non-class pixels mostly outside the contours. \textbf{B} left: deriving a probabilistic shape model from training object masks. right: shape term is low when the model's probability of the evolving shape is high. \textbf{C} left: constructing a location prior around initial position provided by object detection network - probability of correct location shown by color scale from high (yellow) to low (blue). right: location term is low when the center of the evolving object (red point) is deemed probable by location model \textbf{D} Overlap: energy is low when evolving shapes do not overlap. }
	\label{fig:energyTermsExplained}
\end{figure*}

The location model specifies how far away the evolving object centers are allowed to be dislocated from the initial position estimates $\tilde{\omega}^{ctr}_{i}$. A natural choice is an isotropic truncated 2D Gaussian or even uniform probability model, centered on $\tilde{\omega}^{ctr}_{i}$, where the support size (i.e. \{$\omega \in \Omega: \mathcal{P}_{loc}(\omega|\tilde{\omega}^{ctr},c) > 0\}$ may be derived based on the knowledge of the image resolution as well as the average size of class $c$ objects. The orientation model can be used to specify domain-knowledge constraints about implausible object configurations within the scene, e.g.~enforcing objects lying on a flat surface to be either horizontal or vertical, but not 'diagnonal'. Finally, the interaction model plays an important role of discouraging two evolving shapes from occupying the same image space. This is one of the key factors driving the evolution, pushing the contours away from each other and thus reinforcing the incentive generated by the image energy (Eq.~\ref{eq:eimg}) to cover high-probability 'object' image regions with the model shapes. In this work, we penalize all overlap by setting $\log \mathcal{P}_{int}(|M_i \cap M_j|)= -|M_i \cap M_j|$. We note that some applications might require more nuanced approaches.

\subsubsection{General energy formulation}

We may now introduce the proposed energy formulation driving the evolution of $N_o$ contours $C_1,...,C_{N_o}$ identified by their location, orientation, and shape parameters $\pi_i = (\omega^{ctr}_i, \kappa_i, \vec{\alpha}_i)$. Each shape evolves within a class-specific $D_{c_i} \times D_{c_i}$ window centered on $\omega^{ctr}_i$. In analogy to the single-contour Bayesian formulation (Eq.~\ref{eq:bayesianRegions}), we write the aggregate energy as the negative log-probability of the contours given the image evidence, which factors into an image term and the shape-location-orientation prior:

\begin{equation}
\label{eq:multiContourEnergy}
\begin{split}
    &E(C_1,\ldots,C_{N_o}) \equiv -\log \mathcal{P}(\pi_1,\ldots,\pi_{N_o}|I)\\
    &=\underbrace{-\log {\mathcal{P}(I|\pi_1,\ldots,\pi_{N_o})}}_{E_{img}\text{: image term}}
\underbrace{- \log \mathcal{P}(\pi_1,\ldots,\pi_{N_o})}_{E_{pr}\text{:shape/loc./orient.
prior}}
\end{split}
\end{equation}

Figure \ref{fig:energyTermsExplained} explains visually the concept of the various terms that contribute to the aggregate energy of any given shape/orientation parameter configuration.

The image term $E_{img}$ should be understood as the per-pixel cross-entropy between the 'soft labels' resulting from the evolving level-set functions $\phi_i$ and the class posterior probabilities obtained from the semantic segmentation model $\mathcal{P}_{\text{sem}}$. In Section~\ref{sec:imgTerm_singleClass}, we specify the image term for the one-class, multiple-instance case. See Appendix B for the generalization to the multi-class scenario.

First, we turn our attention to the term $E_{sl}$, the analogue of the shape prior from Eq.~\ref{eq:bayesianRegions}. We refer to $E_{sl}$ as the joint shape/location/orientation prior, since it operates on all the evolving shape parameters $(\omega^{ctr}_i, \kappa_i, \vec{\alpha}_i)$ jointly. It should be emphasized that in our setting, the joint prior is fundamentally different than merely applying the single-contour prior to each evolving contour separately due to the inherent conditional dependencies between the object shapes and positions. Indeed, by introducing the shape interaction model $\mathcal{P}_{int}$, we discourage shapes from occupying the same image image space, resulting in a force of 'repulsion' between adjacent model objects, and thus rendering their shapes and positions not independent. To model these interactions in a structured manner, we construct an undirected probabilistic graphical model to serve as the joint shape/location prior. It is defined on the graph $G=(\mathcal{V},\mathcal{E})$ such that the nodes $\mathcal{V}$ correspond to the evolving shapes, whereas the edges $\mathcal{E}$ reflect the spatial neighborhood relations between pairs of objects which are 'close enough' to interact.
The pairwise graphical model can be written as:

\begin{equation}
\mathcal{P}(\pi_1,\ldots,\pi_{N_o})\propto\prod_{\lambda \in
\Lambda}\Psi_\lambda(\pi_\lambda)
\end{equation}

The set of cliques $\Lambda$ consists of the following unary and pairwise potentials:
\begin{itemize}
    \item $\Psi^{shp}_{c_i}(\vec{\alpha}_i) \equiv \mathcal{P}^{c_i}_{shp}(\vec{\alpha}_i)$ is a unary potential which quantifies the prior probability of observing a shape of class $c_i$ given by the coefficient vector $\vec{\alpha}_i$ (cf.~Eq.~\ref{eq:shape_prob_alpha_coef}).
    \item $\Psi^{ori}_{c_i}(\kappa_i) \equiv \mathcal{P}_{rot}(\kappa_i|c_i)$ is a unary potential which corresponds to the prior probability of observing an object of class $c_i$ in an orientation $\kappa_i$
    \item $\Psi^{loc}_i(\omega^{ctr}_i)\equiv \mathcal{P}_{loc}(\omega^{ctr}_i|\tilde{\omega}^{ctr}_i,c_i)$ is a unary potential which quantifies the prior probability of model shape $i$'s position being located at coordinates $\omega^{ctr}_i$ given the prior information $\tilde{\omega}^{ctr}_i,c_i$ from the detector (see previous section)
    \item $\Psi_{ovp}(|H[\phi_{\pi_i}] \cap H[\phi_{\pi_j}]|) \equiv P_{int}(|H[\phi_{\pi_i}] \cap H[\phi_{\pi_j}]|)$ is a pairwise potential which corresponds to the repulsion interaction between neighboring shapes (see previous section)
\end{itemize}

While the unary potentials are just applications of models to portions of the shape parameter vectors $\pi_i$, the pairwise interaction potential deserves a more thorough explanation. First, the notation $\phi_{\pi_i}$ indicates the level-set function $\phi_i$ of evolving shape $i$ obtained from (i) decoding the shape coefficients $\alpha_i$ by the \emph{decoder function} $f_d$ (see Sec.~\ref{sec:generalizing_eigenshape_rep}), and (ii) applying the affine transform $\mathcal{T}_i$ (cf. Eq.~\ref{eq:phi_repr_xform}) composed of a rotation by $\kappa_i$ and translation by $\omega^{ctr}_i$:


\begin{equation}\label{eq:affineTransform}
\phi_{\pi_i} = \mathcal{T}_i(\omega^{ctr}_i, \kappa_i)H[f_d(\vec{\alpha}_i)]
\end{equation}

Moreover, due to the nature of the Heaviside function $H[\cdot]:\mathbb{R} \mapsto \{0,1\}$, the cardinality $|H[\phi_{\pi_i}] \cap H[\phi_{\pi_j}]|$ of the intersection of two shapes can be approximated by their product $\int_{\omega} H[\phi_{\pi_i}] \cdot H[\phi_{\pi_j}]d\omega$ integrated over the image, up to border pixel effects\footnote{In practice, the rotation and non-integer translation applied to the Heaviside function would cause fractional values to appear near the shape border within $\phi_{\pi_i}$}.
 
\begin{figure*}[h!]
\centering
		\includegraphics[width=0.9\textwidth]{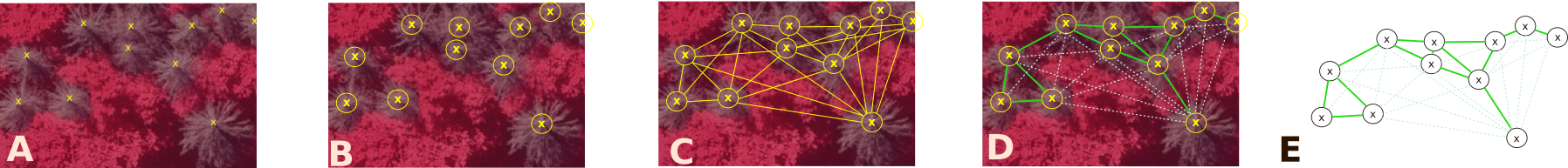}
    \caption{
     Illustration of constructing graph of possible interactions between objects (tree crowns). 
     \textbf{A:} input scene with initially detected object locations 
     \textbf{B:} object centers shown with red circles mark nodes of the interaction graph, green edges are drawn between any two objects that are within interaction distance, based on their mutual location and bounding box sizes. 
     \textbf{C:} abstract representation of resulting interaction graph.
     \textbf{D:} suggested graph with spatially insufficient edges pruned
     \textbf{E:} simplified graph with discarded edges for efficient inference.
     }
	\label{fig:graphConstructionPairwise}
\end{figure*}

Using the above, we can explicitly express the \textit{prior term for the shape/location/orientation}:

\begin{equation}\label{eq:E_prior_terms}
\begin{split}
&E_{pr}(\pi_1,\ldots,\pi_{N_o}) = -\log \mathcal{P}(\pi_1,\ldots,\pi_{N_o})+
\text{const.}=\\
&-\gamma_{shp}\sum_{k=1}^{N_o}\underbrace{\log \Psi^{shp}_{c_k}(\vec{\alpha}_k)}_{\text{shape prior}}\\ 
&- \gamma_{loc}\sum_{k=1}^{N_o}\underbrace{\log
\Psi^{loc}_{k}(\omega^{ctr}_k)}_{\text{location prior}} - \gamma_{ori}\sum_{k=1}^{N_o}\underbrace{\log
\Psi^{ori}_{c_k}(\kappa_k)}_{\text{orientation prior}}\\
&-\gamma_{ovp}\sum_{(k,l) \in \mathcal{E}} \underbrace{\log\Psi_{ovp}(\int_{\omega} H[\phi_{\pi_k}] \cdot H[\phi_{\pi_l}]d\omega)}_{\text{overlap/interaction term}}
\end{split}
\end{equation}
The \textit{coefficients} $\gamma_\cdot$ control the relative influence of the various energy terms. To construct the \textit{graph edges} $\mathcal{E}$, we use: (1) initial object position estimates $\tilde{\omega}^{ctr}_{i}$, (2) bounding boxes $\tilde{M}_i$, and (3) object classes $c_i$ (obtained from the object detection 'oracle'), as input to our framework. 
\jacomment{do we want to simplify this to the distance threshold /  ... derived by maximum size of center-to-center of each tree??}
\jacomment{Based on the size of $\tilde{M}_i$, the center position, and the support region of the location prior $\mathcal{P}_{loc}$ (i.e.~how 'wide' it is), it is possible to estimate a bounding rectangle $B^r_i$ which delineates the extent of the image that could potentially intersect with evolving shape $i$ given the initial conditions and the location prior.}
If distance between trees crown centroids is below a threshold (has high entropy), we connect these nodes (solid lines in Fig.~\ref{fig:graphConstructionPairwise}\,\textbf{D}) and those above the threshold are pruned (dotted lines in Fig.~\ref{fig:graphConstructionPairwise}\,\textbf{D}).
In other words, a graph edge is constructed for every pair of shapes $i,j$ such that their corresponding bounding rectangles $B^r_i,B^r_j$ intersect, thereby simplifying the graph's architecture.  
%
This compact construction of the graph leads to improved scaling behavior
and reduced computational complexity.
Specifically, the number of pairwise interaction terms to evaluate does not scale quadratically with the number of shapes, as in the case of a fully-connected graph. Instead the interactions between nodes are spatially localized and only those nodes/trees deemed 'near enough' to each other are evaluated, thereby increasing computational efficiency.

\subsubsection{Smooth maximum approximation}\label{sec:imgTerm_singleClass}

We will now take a closer look at the image term $E_{img}$ of the aggregate energy from Eq.~\ref{eq:multiContourEnergy}. As stated before, we would like for the image energy to approximate the cross-entropy between 'soft labels' (i.e.~the posterior class distribution from the semantic segmentation model $\mathcal{P}_{\text{sem}}$) and the class assignment distribution obtained from the evolving contours' Heaviside functions $H[\phi_i]$. In the single-contour case, it is natural to treat the value of the Heaviside function as a binary label where $0$ indicates the background class and $1$ indicates the target object class. Extending this to the single-class, multiple-contour case, we note that the evolving contours may overlap at any image element $\omega$, and the 'foreground' class should be assigned if \emph{any} of the contours at $\omega$ produces a 'foreground' value. Using set-theoretic notation, we may write the aggregate Heaviside function $H^{\text{agg}}[\omega|\phi_1,\ldots,\phi_{N_o}]$ over the $N_o$ evolving contours at an image element $\omega$ as:

\begin{equation}
    H^{\text{agg}}[\omega|\phi_1,\ldots,\phi_{N_o}] = \bigcup_{i} H[\phi_i](\omega)
\end{equation}

Translating this from the Boolean setting to the realm of real numbers, we apply the $\max$ operator to implement the set-theoretic union, to ensure normalization of $H^{\text{agg}}$ between $[0;1]$. Thus far, the tacit assumption has been that the employed Heaviside function is 'perfect', i.e.~it produces binary output from $\{0,1\}$. However, since the true Heaviside function is neither continuous nor differentiable, in practice it would not be amenable to gradient-based optimization methods. As we aim for a framework which is readily optimizable with standard methods, we utilize a smooth approximation $\tilde{H}(\cdot)$ to $H$\comment{, where the parameter $\delta$ controls the 'steepness' of the slope marking the transition from the 'low' to 'high' state}. Therefore, the image of $\tilde{H}(\cdot):\mathbbm{R} \mapsto (0;1)$ is now the real interval $(0;1)$ instead of the binary states $\{0,1\}$. Moreover, we require a smooth approximation $\tilde{S}(\cdot, \ldots,\cdot)$ for the non-differentiable $\max$ operator as well, to ensure existence of gradients. Based on the newly introduced smooth approximations, we may define the final single-class image energy term as ($c_b,c_f$ indicate, respectively, the background and foreground class):
\begin{equation}
\begin{split}
    &E_{img} = -\int_{\omega} \tilde{S}(\tilde{H}(\phi_1)[\omega],\ldots,\tilde{H}(\phi_{N_o})[\omega]) \log \mathcal{P}_{\text{sem}}(c_f|\omega)\\
    &+(1-\tilde{S}(\tilde{H}(\phi_1)[\omega],\ldots,\tilde{H}(\phi_{N_o})[\omega])) \log \mathcal{P}_{\text{sem}}( c_b | \omega)d\omega
\end{split}
\end{equation}

In principle, any smooth approximation $\tilde{H}$ to the Heaviside function may be used, (e.g.~logistic function, or arc tangent). Similarly, there are several choices for the smooth maximum approximation (see \ref{appendix:smoothMax} for details).

\subsubsection{Initializing the shapes}\label{sec:initializeShapes}

\begin{figure}[h!]
\centering
	\subfloat[initial detection\label{1a}]{%
       \includegraphics[width=0.32\linewidth]{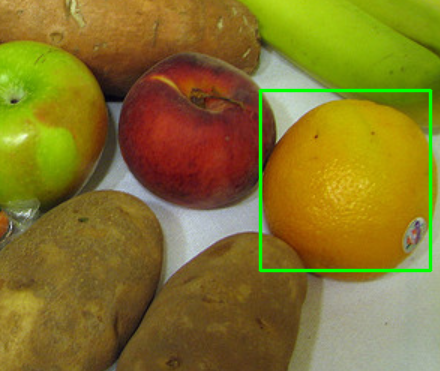}}
    \hfill
  \subfloat[intersection\label{1b}]{%
       \includegraphics[width=0.32\linewidth]{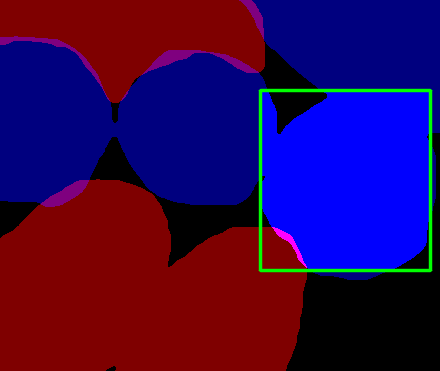}}
    \hfill
    \subfloat[projection\label{1c}]{%
       \includegraphics[width=0.32\linewidth]{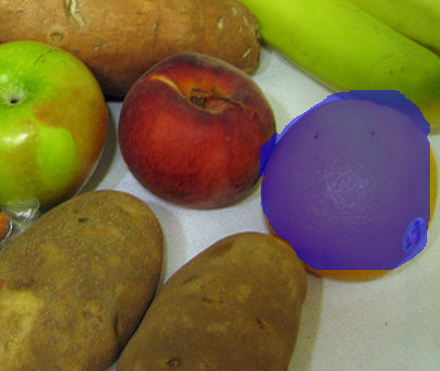}}
	\caption{Three stages of initializing shape coefficients for a detection. (a) bounding box of initial detection received as input. (b): intersection of detected bounding box with semantic segmentation map of appropriate class. (c): intersection from center projected onto the principal components of the shape model, decoded back into image space.}
	\label{fig:shapeCoeffInitialize}
\end{figure}

Given the initial object center positions $\tilde{\omega}^{ctr}_{i}$, detected classes $c_i$ and bounding boxes $\tilde{M}_i$, it is necessary to translate them into the initial parameter set $\pi_1^0,\ldots,\pi_{N_o}^0$ describing the evolving shapes. This is an important part of our approach which avoids the need to do costly random restarts of the optimization. To that end, we compute an intersection of each bounding box $\tilde{M}_i$ with the semantic segmentation map of class $c_i$ defined as: $M^{sem}_i[\omega]=\delta(\arg \max_{c_k} \mathcal{P}_{\text{sem}}(c_k|\omega),c_i)$, where $\delta$ represents the Kronecker delta. In other words, we obtain a binary image $M^{init}_i$ such that a pixel $\omega$ has a 'set' value iff it lies inside the input bounding box $\tilde{M}_i$ \emph{and} the dominant class in the semantic segmentation map ${P}_{\text{sem}}$ at $\omega$ is the detected class $c_i$. Furthermore, the image $M^{init}_i$ is centered at $\tilde{\omega}^{ctr}_{i}$ and cropped to size $D_{c_i} \times D_{c_i}$. The next step depends on the presence of orientation in the affine transform. As our application domain features a target object that does not posess a well-defined 'standard' orientation (i.e.~tree crowns in the nadir view), we consider the case when the affine transform is based on translation only. In this case, initializing our model parameters $\pi_i$ amounts to simply obtaining the shape representation of $M^{init}_i$ in terms of the learned shape model, i.e.~projecting $M^{init}_i$ onto the principal components obtained from the kernel PCA procedure (see Eq.~\ref{eq:projectingShapeCoefs} and Fig.~\ref{fig:shapeCoeffInitialize}). This yields the initial shape coefficients $\vec{\alpha}^0_i$, which, together with the initial center positions $\tilde{\omega}^{ctr}_{i}$, fully describe the state of the shape evolution. We propose some extensions to cases which include the rotation/orientation parameter in \ref{sec:appendixInitParam}.

\subsection{Deep shape models}\label{sec:deepShapeModels}

\begin{figure}[h!]
\centering
		\includegraphics[width=0.9\columnwidth]{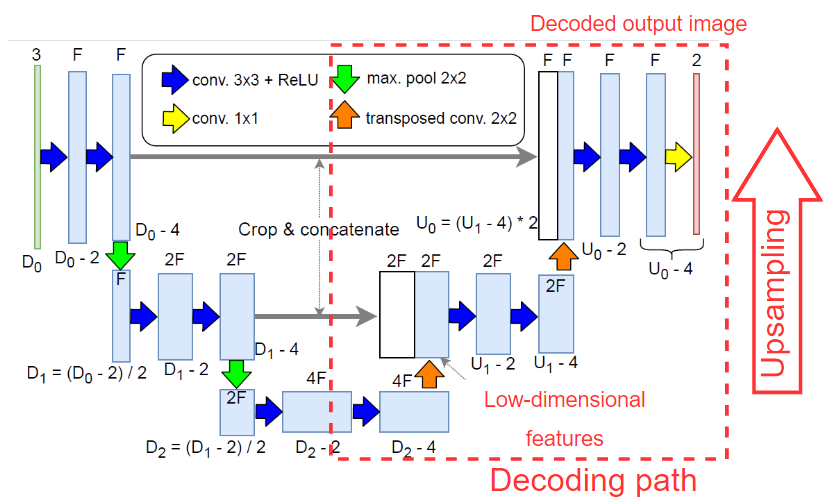}
	\caption{Example of a fully convolutional network with symmetrical encoding \& decoding paths: the U-net \citep{RFB15a}. The \textit{decoding} path (using transposed convolutions) gradually \textit{up-samples} low-dimensional features to produces an image of the original size (full-size).}
	\label{fig:unetDecoderPath}
\end{figure}

In this section, we propose a new class of non-linear, trainable shape models based on CNN architectures akin to the 'decoder' networks that are found in many areas of the CNN landscape. A prime example is that of fully convolutional network (FCN) based methods that are currently the state-of-the-art in semantic image segmentation~\citep{9356353}. Most of these methods share the \emph{encoder-decoder} design, which partitions the network into 2 symmetrical, opposing paths: the \emph{encoder} transforms the input image into a low-dimensional representation based on a sequence of convolution and pooling operators, while the \emph{decoder} up-samples the compact feature representation back into a full-sized image, again through the use of convolutions and transposed/fractional convolutions~\citep{https://doi.org/10.48550/arxiv.1603.07285} in place of the pooling layers (see Fig.~\ref{fig:unetDecoderPath} for an example).

Decoder architectures are often applied as \textit{generators} within the \textit{generative adversarial network (GAN) framework}. Indeed, for the domain of image data, the role of the decoder is to transform a low-dimensional \emph{latent variable} vector into the generated image. In case of GANs, the network training is driven by a two-player, zero-sum \emph{game} between the generator/decoder and the \emph{discriminator}. The goal of the \textit{discriminator} is to distinguish between real and 'fake' (generated) data instances, whereas the goal of the generator is to deceive the discriminator by generating samples which are indistinguishable from the original training data distribution. Thus, the generator $G_z$ learns to map random latent vectors $\vec{z}$ sampled from the latent distribution $\mathcal{P}_z$ to the space of output images, such that the distribution of $G_z(z)$ approximates the distribution of the original training imagery. It should be noted that the adversarial training occurs \emph{without an explicit pairing} between the latent vectors and the images they should generate. However, we show that if the topology of the latent space is already known and explicit pairs $(z,G(z))$ are available, the generator network can be re-used and trained in a supervised manner, resulting in a quality shape model for our instance segmentation framework.


First, we will take a closer look at one of the first fully convolutional GAN architectures, proposed by~\cite{DBLP:journals/corr/RadfordMC15}. The design of the generator network 
included some architectural innovations such as: (i) refraining from the use of fully connected layers, (ii) using fractional-stride/transposed convolutions for the up-sampling operation, which affords trainable parameters in upsampling, and (iii) applying batch-normalization layers for better stability of the training process. The vector of $c$ latent variables is projected (via a dense layer) onto an initially small spatial extent $D^0 \times D^0$ (e.g.~4x4 pixels) with an initial number $N_{conv}^0$ of convolutional filters. In each subsequent step, the number of filters is halved, while the spatial extent is doubled in each dimension, using transposed convolutions with filter size $d_f$. The upsampling is continued until the desired output image size $D^{out} \times D^{out}$ is reached. The final output up-convolution layer contains a reduced number of filters, corresponding to the number of output channels in the generated image (e.g.~3 for RGB and 1 for binary images). Each 'intermediate' convolution layer (except the final output) is followed by a batch normalization operation and a leaky ReLU activation, whereas the final output layer utilizes a hyperbolic tangent activation.

\begin{figure}[h!]
\centering
		\includegraphics[width=0.9\columnwidth]{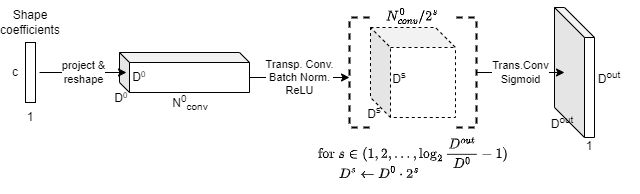}
	\caption{Architecture of proposed deep shape model, generalizing and parameterizing the DCGAN generator \citep{DBLP:journals/corr/RadfordMC15}.  The vector of $c$ latent variables is projected on a $D^0 \times D^0$ spatial extent with $N^0_{conv}$ convolutional filters. In each step, the spatial dimension is doubled while the filter depth is halved, until the desired image size of $D^{out}$ is attained.}
	\label{fig:genericDecoderShapeModel}
\end{figure}

Based on this reference architecture, we define a family of decoder networks $\mathcal{F}_d=\{f_d(\cdot|c,d_f,N_{conv}^0,D^0,D^{out})\}$, parameterized by the number of latent variables $c$, the initial and target image sizes $D^0,D^{out}$, as well as the initial filter count and filter size of convolution layers $N_{conv}^0,d_f$. For convenience, we assume that $D^{out} / D^0=2^u$, and that $N_{conv}^0$ is a multiple of $2^u$ (see Fig.~\ref{fig:genericDecoderShapeModel} for an illustration of this idea). Given a parametrization, a specific $f_d \in \mathcal{F}_d$ can be trained on pairs $\{(\vec{\alpha_i}, M_i)\}$ of corresponding $c$-length shape coefficient vectors and the masks which the network should produce in response to these 'stimuli', guided by e.g.~a per-pixel binary cross-entropy objective comparing the output of the network's final layer to the target mask. The set of trainable parameters includes the weights of the initial dense layers projecting the latent variable vector (i.e.~shape coefficients) onto the first convolutional layer, as well as the parameters of the transposed convolution kernels. Training can be performed using standard gradient-based methods, back-propagating the error through the intermediate layers to the weights. Once the kernel PCA shape coefficient model (see Sec.~
\ref{sec:generalizing_eigenshape_rep}) is constructed, it is easy to project any training mask $M_k$ onto the principal components as per Eq.~\ref{eq:projectingShapeCoefs}, thus obtaining the \emph{latent code - decoded image} pairs. By training the network, we encode a non-linear shape model implicitly in the network weights, departing from the explicit additive model of (linear) eigenshapes, where the eigenvectors of shape variations are easily obtainable.





\begin{figure*}[ht]
  \centering
    \centering
  \subfloat[dead tree crowns (CIR image)\label{gt_1a}]{%
       \includegraphics[width=0.32\linewidth]{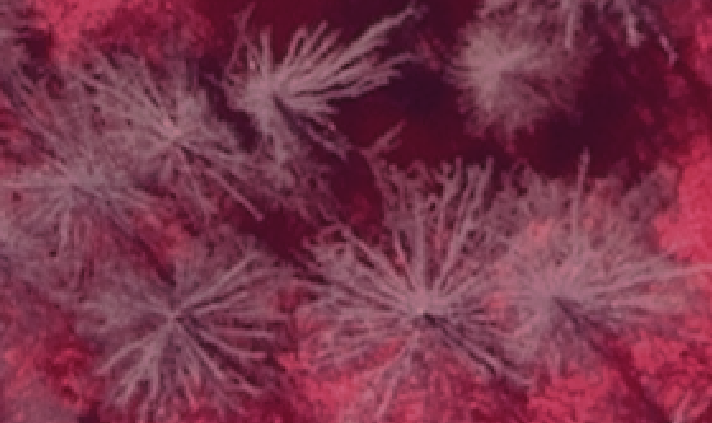}}
    \hfill
  \subfloat[manually drawn tree crown contours\label{gt_1b}]{%
        \includegraphics[width=0.32\linewidth]{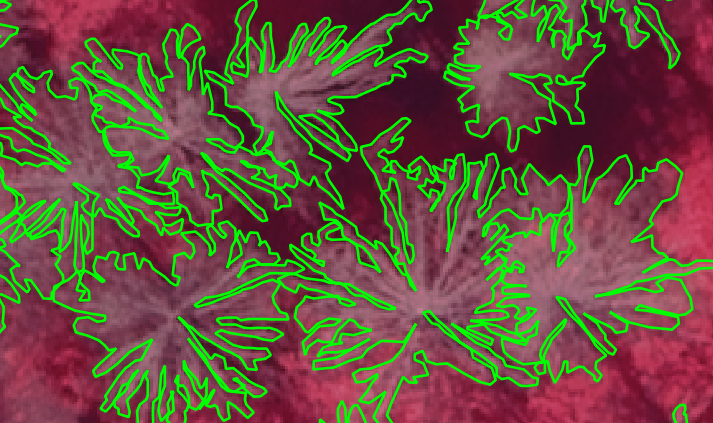}}
  \hfill
  \subfloat[training data masks for U-net]{%
        \includegraphics[width=0.092\linewidth]{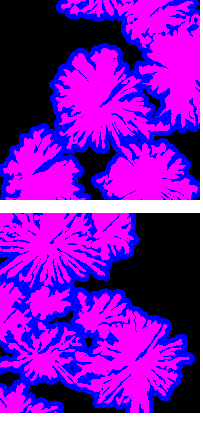}}
  \hfill
  \subfloat[binary masks learnt - use to train shape models\label{gt_1c}]{%
        \includegraphics[width=0.19\linewidth]{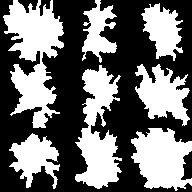}}
  \caption{Input imagery and derived ground truth data for training. 
    Shown: (a) sample dead tree crowns visible on color infrared imagery, 
    (b) the same scene with manually marked contours of each tree crown, 
    (c) input training mask for the U-net. The positive and negative class pixels shown respectively in \textcolor{magenta}{magenta} and \textcolor{blue}{blue}, 
    (d) sample binary masks of individual dead tree crowns obtained from manual labeling of imagery as in (b), used as input for training the shape models (eigenshape and deep shape/dcgan model).}
\end{figure*}\label{fig:trainingData}

\section{Experiments}
We validate our proposed method in two flavors: with vanilla eigenshape and our deep shape model, compared with Mask R-CNN~\citep{8237584} and K-net~\citep{zhang2021knet} as baselines.
In these experiments, we address a critical and challenging problem from the field of environmental science and precision forestry: real-world object segmentation of remote sensing imagery, specifically the fine instance segmentation of individual tree crowns in high-resolution aerial images.

\subsection{Problem setting: Image segmentation in precision forestry}\label{sec:problemSetting} \jacomment{: Forestry? smth else? else leave as is}

Forest ecosystems are the most species-rich ecosystems on Earth and are the key components in ecological processes such as wood production, drinking water supply, carbon sequestration, and biodiversity
preservation~\citep{watsonExceptionalValueIntact2018}. One of the crucial indicators of forest health is the quantity and distribution of dead wood, as e.g.~in temperate forests, up to
one-third of all species depend on it during their life
cycle~\citep{Mller2010ARO}. Moreover, it is not just the amount of deadwood
that matters for the conservation of biodiversity~\citep{Seibold2018}; also its
quality, as determined by parameters such as the tree species, decay stage etc., is decisive for the conservation of biodiversity. Therefore, accurate mapping of deadwood can be seen as critical for monitoring the effects of forest disturbances (e.g.~insect outbreaks) which have been increasing in frequency and severity due to global change~\citep{Seidl2017}.

Remote sensing techniques based on aerial imaging offer a unique possibility of large-scale deadwood mapping, owing to the fact that that dead and
diseased vegetation produces a distinct reflectance signature in the near-infrared band~\citep{Jensen06}. Moreover, the progress in imaging hardware and the rising availability of unmanned aerial vehicles in civil applications~\citep{Cummings2017} make aerial imaging an appealing and cost-efficient alternative to manual field measurements. Specifically, we tackle the problem of delineating precise, fine-detailed, contours of individual dead tree crowns in very high resolution (10 cm ground sampling distance) aerial imagery as the testbed for our methods. The tree crown shapes, visible with unprecedented detail within the imagery, make for an appropriate setting for evaluating the methods' capability to reconstruct fine contour elements. While the general task of delineating individual dead trees from imagery has been attempted before (e.g.~\citep{isprs-annals-II-3-W4-181-2015,f11080801}), the authors are not aware of any study based on imagery with similarly high resolution. Concurrently, several recent works have extended dead-tree and canopy-dieback detection to multi-biome, multi-resolution, or multi-sensor settings, e.g. combining transformer-based segmentation across globally distributed aerial imagery ~\citep{MOHRING2025100104}, comparing deep learning architectures across ecologically distinct forest sites~\citep{10.1093/forestry/cpag044}, fusing self-attention U-Nets with watershed-based post-processing ~\citep{RAHMAN2025104851}, and integrating UAV RGB, multispectral, and LiDAR data for standing dead tree classification ~\citep{https://doi.org/10.1002/rse2.70070}. These studies, however, either operate at coarser spatial resolutions or do not specifically target the reconstruction of fine-grained, individual crown contours, which remains the focus of the present work.

\subsection{Imagery acquisition}

\jacomment{reorder points in paper, emphasise why we care about THIS data / why helpful application to forestry}
The color infrared imagery was acquired during a flight campaign carried out in June 2017 within the Bavarian Forest National Park, situated in South-Eastern Germany ($49^\circ 3' 19''$
N, $13^\circ 12' 9''$ E). From
1988 to 2010, a total of 5800 ha of the Norway spruce stands died off because
of a bark beetle (\emph{Ips typographus}) infestation~\citep{Lausch2012}. The study area was located inside the region of the park strongly affected by the outbreak. The mean above-ground flight height was ca.~2879 m, resulting in a pixel
resolution of 10 cm on the ground. The acquired imagery was radiometrically corrected and consisted of 3 spectral bands: near
infrared (spectral range 808-882 nm), red (619-651 nm) and green (525-585 nm).

\begin{figure*}[ht]
  \centering
       \includegraphics[width=\linewidth]{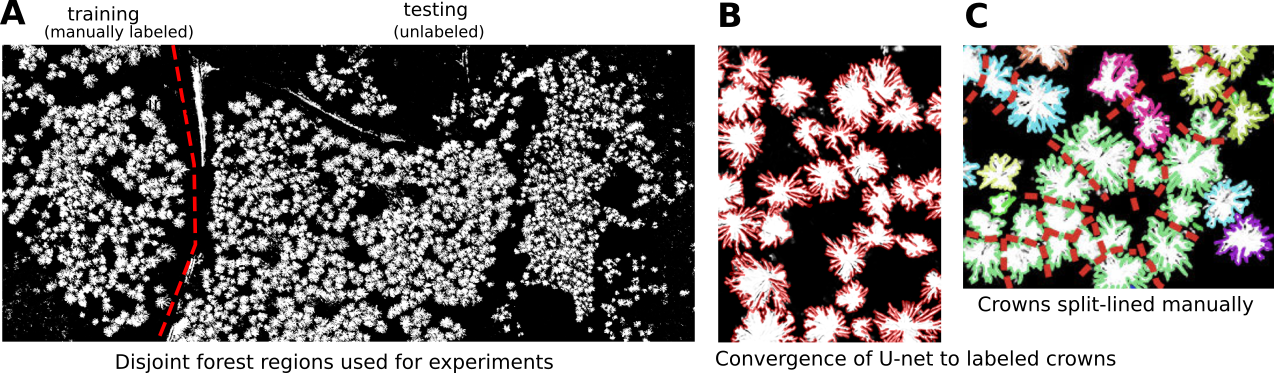}
  \caption{(A) Semantic segmentation of source aerial CIR image with marked partition between disjoint training and test areas in the Bavarian Forest National Park. (B) Convergence of U-net based semantic segmentation to manual contours on training data (\textcolor{red}{red} polygons). Pixel intensity indicates dead tree crown probability. (C) Semi-automatic ground truth derivation by splitting the semantic segmentation result polygons into individual tree masks using manually drawn split lines (dashed \textcolor{red}{red} lines). Crown polygon outline colors indicate the connected component in the graph of high-probability dead tree pixels from semantic segmentation, as determined by N8-connectivity.   
  }
  \label{fig:trainTestSplit_unetConvergence}
\end{figure*}

\subsection{Datasets and ground truth}\label{sec:groundTruth}

\jacomment{------- strengthen this statement about the cost of labels -- see papers semi-sup and efficient gibbs - e.g. junction tree to save these costs and acquire good GT approxes -- save costly manual labelings of contours and inconsistencies across multiple labelers -- all data conditions of train/test sets identical ... same high res aerial image with 200 labels ---}

Ground truth data is based on 200 contour polygons of dead tree crowns, manually marked on the target color infrared image (see Fig.~\ref{gt_1a},~\ref{gt_1b}) by a single analyst within a short time span, in order to maintain homogeneity of the labels and avoid subjective differences in delineating fine contour details. For experiments involving the comparison of our entire pipeline to baseline methods (K-net, Mask R-CNN), the dataset was partitioned into training and validation sets in approximately a 1:3 ratio (see Fig.~\ref{fig:trainTestSplit_unetConvergence}\,\textbf{A}).  To train shape models, the marked contours were converted into binary shape masks and used directly as input training data (see Fig.~\ref{gt_1c}). As for the semantic and instance segmentation networks, we cropped patches from the source imagery such that every marked polygon was entirely visible in at least one patch. Note that not all dead trees visible in each cropped patch were annotated as ground truth. To avoid biases in the training, we used network-specific mitigation strategies (see Sec.~\ref{sec:partialModels}).

Given that the manual effort required to delineate individual tree crowns is quite high (cf.~Fig.~\ref{gt_1b}), and on the other hand that we wanted to validate our study on a larger data sample, we decided to follow a semi-automatic strategy for obtaining more labeled tree crown polygons. Our strategy is based on the observation that the semantic segmentation network (U-net) was able to converge to a high-fidelity representation of the manually marked polygons on the training area, capturing fine contour details (see Fig.~\ref{fig:trainTestSplit_unetConvergence}\,\textbf{B}). Also, it is much easier and less time-consuming to mark \emph{boundaries} between neighboring tree polygons by drawing linear features (Fig.~\ref{fig:trainTestSplit_unetConvergence}\,\textbf{C}) than it is to delineate each polygon's detailed contour. Based on these assumptions, we applied the trained U-net to a region of the color infrared image immediately adjacent to the training area, covering a larger part of a contiguous forest stand, separated from the training part by a dirt road (see Fig.~\ref{fig:trainTestSplit_unetConvergence}\,\textbf{A}). We marked a total of 528 split lines on the obtained semantic segmentation map (thresholded on 0.5 target class probability), resulting in individual dead tree masks. After visual inspection, we selected a total of 759 of these binary masks to form our additional validation dataset. This semi-automatic data was used only for comparing the impact of the Deep Shape Model versus the Eigenshape baseline.

\begin{figure*}[h!]
\centering
	\subfloat[Weighted IoU\label{1a_singleTree}]{%
       \includegraphics[height=100px]{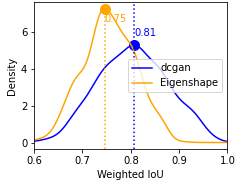}}
    \hfill
  \subfloat[Target shape\label{1b_singleTree}]{%
       \includegraphics[height=100px]{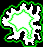}}
    \hfill
    \subfloat[DCGAN result\label{1c_singleTree}]{%
       \includegraphics[height=100px]{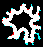}}
     \hfill
    \subfloat[Eigenshape result\label{1d_singleTree}]{%
       \includegraphics[height=100px]{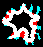}}
	\caption{Results of \emph{Experiment 1}. (a) Distribution of weighted intersection-over-union metric for single tree reconstruction by the eigenshape and DCGAN-based shape models. (b) target shape mask shown in white, 3-pixel band around shape boundary delineated in green. (c)-(d) visualization of a sample reconstruction result by respectively the DCGAN and the eigenshape models within the 3-pixel band. \colorbox{black}{\color{white}White}, \colorbox{black}{\color{cyan}cyan}, \colorbox{black}{\color{red}red} pixels denote respectively correct reconstruction, omission error (ground-truth only pixel), commission error (model-only pixel).}
	\label{fig:singleTreeResultSummary}
\end{figure*}

\subsection{Training the partial models}\label{sec:partialModels}

This section describes the specific training procedures and settings for all partial models utilized in our pipeline.

\subsubsection{\emph{Semantic segmentation network}}
A 3-layer instance of the U-net~\citep{RFB15a} (see Fig.~\ref{fig:unetDecoderPath} for architecture) was trained on cropped 200x200 pixel patches centered around each of the labeled tree crown polygons from the training area. Pixels inside and outside the polygons constituted respectively positive and negative class examples. Since not all tree crowns visible in the target imagery were labeled, we constrained the negative class range to a band of 5 pixels around the marked polygons, thus avoiding errors of mis-interpreting unmarked dead trees within the images as the negative class (background) (see Fig.~\ref{gt_1c}). We used
the Adam algorithm with standard metaparameters ($\alpha=0.001,
\beta_1=0.9,\beta_2=0.999$) to perform
stochastic gradient optimization of a binary logistic objective. The dropout rate was 50 \%, while the training minibatch size was set to 15. The \emph{tensorflow} implementation was adapted from the publicly available Github repository~\citep{akeret2017radio}.

\subsubsection{\emph{Object detection network}}

 An instance of Mask R-CNN with a resnet-50 backbone trained over 100 epochs on 256x256 pixel patches of aerial imagery, with individual tree crown masks as the ground truth labels. Image augmentation was applied in the form of horizontal and vertical flipping, as well as random rotation by multiples of 15 degrees in the range of $[15;345]$. To mitigate the training bias of trees visible in the training imagery which were not labeled as ground truth, we overwrote the corresponding image pixels of these objects with the mean background color.

 \subsubsection{\emph{Deep shape model}}

 To train the deep shape model network (Sec.~\ref{sec:deepShapeModels}), we first obtained the shape coefficient representation of training binary masks representing the dead tree polygons via Principal Component Analysis of the training shape matrix as described in Sec.~\ref{sec:generalizing_eigenshape_rep}, Eq.~\ref{eq:shapeCoefTrain}. We then optimized the network parameters in a supervised setting, learning to decode the shape coefficients into associated images under the binary cross-entropy criterion. Once again, the Adam optimizer was used, with a learning rate of $1e-4$ and standard values for remaining metaparameters. The optimization was carried out over 1000 epochs with a minibatch size of 64. The decoder model was implemented based on the\emph{keras} framework. We created 3 versions of the network, using 32, 64, and 128 shape coefficients as input. Aside from the number of inputs, the structure of the network was kept constant as $f_d(\cdot|c=\cdot,d_f=3,N_{conv}^0=256,D^0=6,D^{out}=96)$. The network operates on an output image of 96x96 pixels, which limits the maximum size of a tree crown to 9.6 x 9.6 meters based on the ground sampling distance of the CIR imagery.


\subsection{Evaluation criteria}\label{sec:evalCriteria}

We evaluate all instance segmentation results on two levels:
\begin{itemize}
    \item \emph{Pixel level}: comparison between the pixel masks of the matched detected vs. ground truth polygons using the usual intersection-over-union (IoU) metric and a weighted IoU version (wIoU) akin to~\cite{DBLP:journals/corr/abs-2107-09858}, where we only consider pixels within an $\epsilon_d$-wide band around the reference polygon's contour (we set $\epsilon_d=3$). The wIoU metric reflects the quality of reconstructing fine shape details around the contour.
    \item \emph{Object level}: standard object detection metrics of accuracy, precision, and recall, based on pairs of matched detected/ground truth polygons such that the pair's IoU value is at least $\text{IoU}_{min} = 0.7$. Pairs with an IoU below the threshold are considered unmatched.
\end{itemize}

\begin{figure}[htbp]
  \centering
  \includegraphics[width=\linewidth]{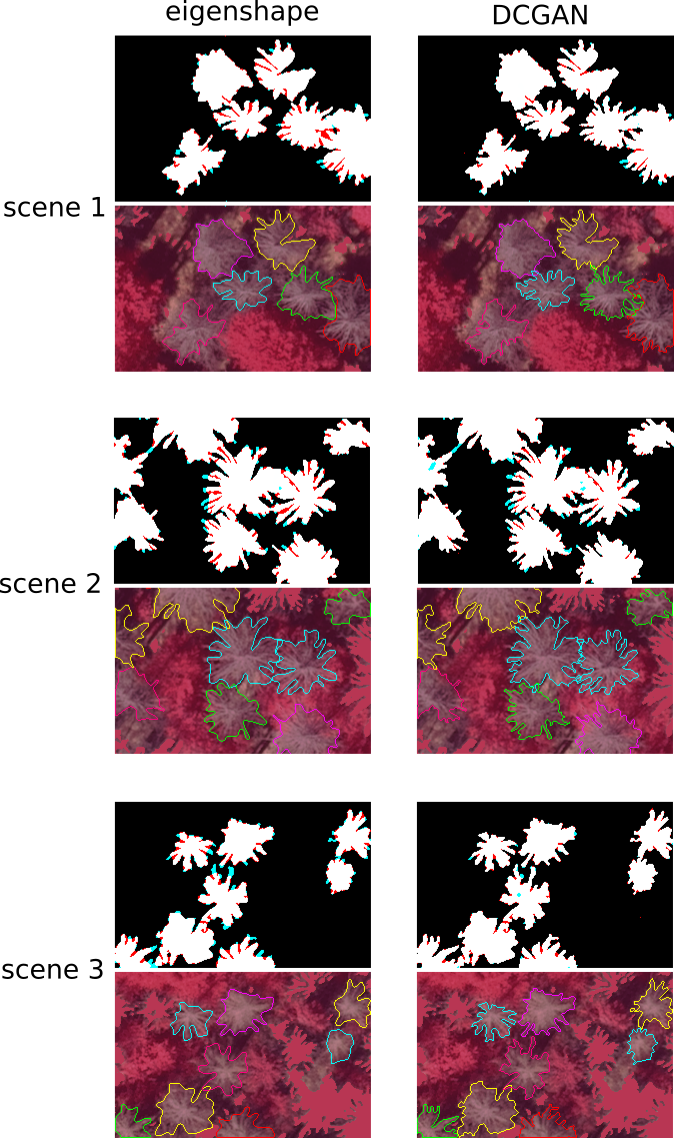}
  \caption{Visual comparison of segmented contour quality on 3 sample test scenes between two instantiations of our framework: Eigenshape (left) and DCGAN/deep shape model (right). For each scene, the top row compares the ground-truth (GT) binary masks to segmented polygons. \colorbox{black}{\color{white}White}, \colorbox{black}{\color{cyan}cyan}, \colorbox{black}{\color{red}red} pixels denote respectively agreement, omission error (ground-truth only pixel), commission error (model-only pixel). The bottom row per scene shows segmented contours over the original CIR image (random outline colors intended to help discern between overlapping shapes). The number of commission \textcolor{red}{error pixels} is significantly reduced by the deep shape model in all scenes, while the number of \textcolor{cyan}{omitted pixels} is maintained or reduced (scene 3). Contours produced by deep shape model are more detailed and align better with ground truth despite the same number of shape coefficients. }
  \label{fig:sampleSegmentedResults}
\end{figure}

\subsection{Implementation notes}

The entire processing pipeline for the optimization of the energy criterion from Eq.~\ref{eq:multiContourEnergy} is done within the tensorflow framework to enable efficient computations on GPUs. The optimization is carried out using tensorflow's interface to the SciPy optimizer, which implements the limited memory BFGS algorithm. In order to include the per-shape affine transform variables as per Eq.~\ref{eq:affineTransform} into the processing pipeline, we used an implementation of the spatial transformer network~\citep{NIPS2015_33ceb07b} available on GitHub~\citep{kevinzakka/spatial-transformer-network}. We used the publicly available Github implementation of Mask R-CNN from the MM Detection Toolbox~\citep{mmdetection}. The second baseline, K-net, was trained using software officially released in conjunction with the paper~\citep{zhang2021knet}. The coefficients $\gamma_{shp}, \gamma_{ovp}, \gamma_{loc}$, controlling the relative impact of the various energy terms from Eq.~\ref{eq:E_prior_terms}, were set empirically at the start of the experiment and were left unaltered for all computations. As discussed in Sec.~\ref{sec:evolvingMultipleShapes}, we used the knowledge of the image ground sampling distance and maximum tree crown dimensions to define the size of the window inside which the shape evolution takes place as 96x96 pixels. All experiments were performed on a NVIDIA GeForce RTX 2080 Ti GPU, with NVIDIA CUDA Toolkit version 11.6. 


\subsection{Results}

In this section, we evaluate the performance of our instance segmentation model on the data described above.


\subsubsection{\emph{Experiment 1:} shape models - individual objects}

We first compared the expressiveness of the decoder-based shape model versus the classic eigenshape baseline, in the simplest setting of reconstructing single objects (tree crowns) with a known center position. The influence of multiple overlapping shapes in the image as well as of finding the actual object center were thus removed, leaving only the task of finding the shape coefficients which lead to the optimal reconstruction of the given object mask under the model. To that end, we used the manually labeled tree crown polygons in Dataset A to train an instance of the decoder shape model as described above (Sec.~\ref{sec:partialModels}). A matching eigenshape model was extracted from the same set of training masks, with the same parameters of $c=64$ eigenmodes corresponding to the latent variables. Both models were then applied to the 759 individual tree crown polygons from Dataset C, unseen during training. The average weighted IoU values for the decoder-based and eigenshape models were respectively 0.80 and 0.75, whereas the corresponding unweighted IoU values were measured as 0.88 and 0.84. The distribution of weighted IoU and a sample result is depicted in Fig.~\ref{fig:singleTreeResultSummary}.\\
\jacomment{FOR ALL RESULTS REPORTED: will simplify and more clearly explain trends, e.g. 36$\%$ better than eigenshape ..}\\ 

\begin{figure*}[ht]
  \centering
    \centering
       \includegraphics[width=0.13\linewidth]{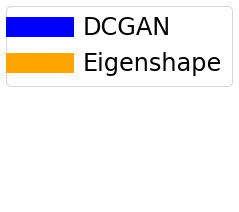}
  \subfloat[]{%
       \includegraphics[width=0.80\linewidth]{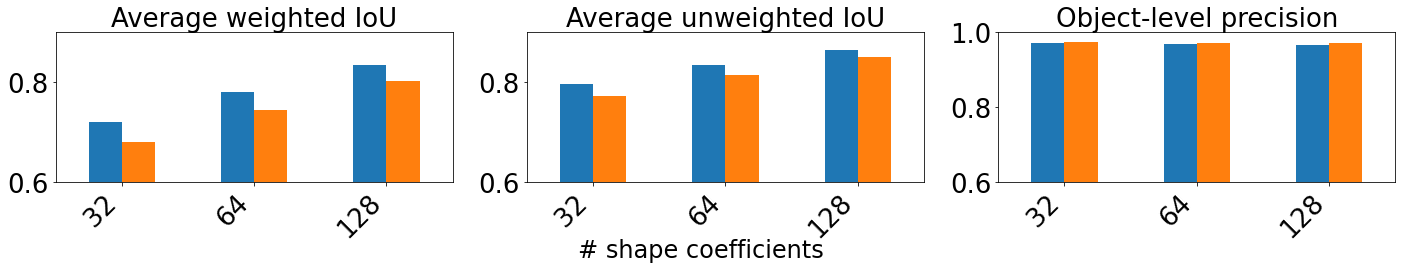}}
 
    \subfloat[\label{fig:dcgan_eig_comparison_bottomRow_part1}]{%
       \includegraphics[width=0.50\linewidth]{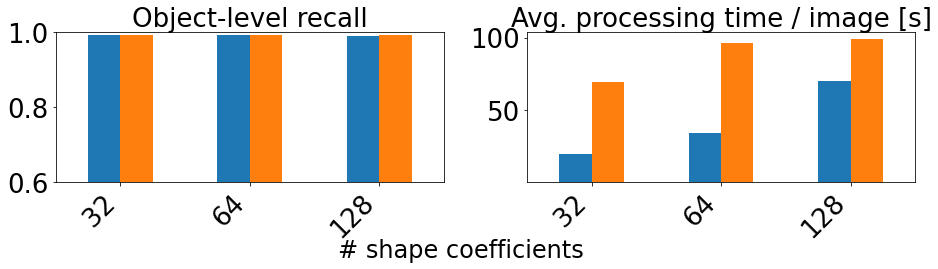}}
    \subfloat[]{%
       \includegraphics[width=0.50\linewidth]{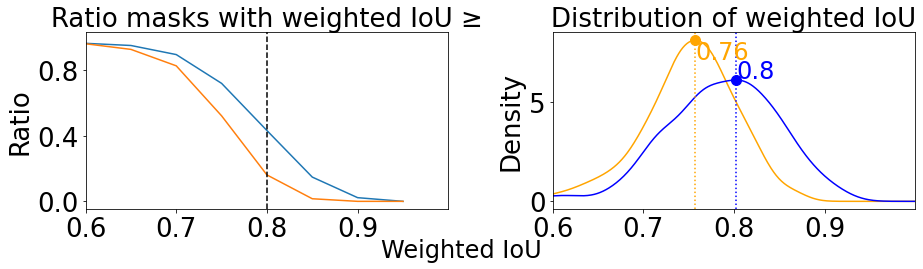}}
  \caption{Quantitative comparison for various metrics on test data between 2 instantiations of our framework: \textcolor{blue}{deep shape model/DCGAN}-based and \textcolor{orange}{eigenshape}-based. (a)-(b) The deep shape model maintains an advantage on the contour-weighted IoU metric at all model complexities, scoring 4 pp higher for 32 and 3.2 pp higher for 128 shape coefficients. The advantage in unweighted IoU ranges from 2.5 to 1.5 pp, indicating the improvement is focused around the contours. Object-level metrics are similar, since most contours are detected to a quality exceeding the 0.7 IoU cutoff by both methods. Average processing time shows that the deep shape model is more performant for all model complexities, although the gap in processing time narrows as the complexity increases. (c) comparison of weighted IoU distributions; left: ratio of test polygons achieving a weighted IoU of $x$ or more (weighted IoU cutoff is $x$ axis). For the DCGAN-based deep shape model, about 43 \% of result polygons have a weighted IoU of 0.8 or better (black dashed line shows cutoff point), vs.~16 \% for Eigenshape. Right: weighted IoU distributions (32 shape coefficients) show a clear advantage of the DCGAN model in the frequency of weighted IoU $\in [0.8;1.0)$}
  \label{fig:chartsCmpEigDcgan_multiShape}
\end{figure*}


\subsubsection{\emph{Experiment 2:} shape models - multiple objects}

In the second experiment, we assessed the impact of the two shape models (eigenshape vs.~deep shape model) on the segmentation quality in the presence of multiple objects. For this experiment, all 200 manually labeled polygons were used in a training capacity. We applied the trained Mask R-CNN to the semi-automatic test area (see Fig.~\ref{fig:trainTestSplit_unetConvergence}\,\textbf{A}) and used the obtained dead tree crown center positions as initial values for our multiple shape evolution process (Sec.~\ref{sec:evolvingMultipleShapes}). The multi-shape evolution was separately carried out for the eigenshape and the deep shape model, yielding two sets of dead tree mask results. Both methods used the same semantic segmentation map from the U-net as its underlying input. That combined with using the same object initializations allows us to attribute any differences in performance to the shape models themselves. The shape evolution was initialized deterministically as per Sec.~\ref{sec:initializeShapes} without random restarts. For each set of results (both shape models and both baselines), we performed a per-instance matching with respect to the ground truth polygons (see Sec.~\ref{sec:groundTruth}), considering a pair of polygons as matched if their intersection-over-union (IoU) metric was at least 0.7. We then applied the object-level and pixel-level metrics enumerated in Sec.~\ref{sec:evalCriteria}. The aggregated results are depicted in Fig.~\ref{fig:chartsCmpEigDcgan_multiShape}, whereas Fig.~\ref{fig:sampleSegmentedResults} highlights differences in quality across the various methods for sample scenes.

For all levels of shape coefficient counts, our proposed deep shape model outperforms the eigenshape counterpart in terms of both weighted and unweighted IoU, while the object-level precision and recall remains on a similar level. Considering the model version trained with the highest degree of rotation augmentation (every 15 degrees), the gains were 4.0 pp (percentage points), 3.7 pp and 3.1 pp in the weighted case, and  2.3 pp, 2.0 pp, and 1.4 pp in the unweighted case respectively for 32, 64, and 128 shape coefficients. The weighted IoU improvements were nearly twice as large as the regular IoU gains, which confirms that fine contour details of the shapes benefited the most from the new model. It is interesting to note that the contours produced by the eigenshape model seem to be more simplified compared to their Deep Shape Model counterparts (see Fig.~\ref{fig:sampleSegmentedResults}). We hypothesize that this phenomenon may be attributed to the different way both models make use of the training set. The deep shape model attempts to learn a decoder network that is able to reconstruct \emph{each individual} training example in a supervised manner, whereas the eigenshape model applies averaging and singular value decomposition to the entire dataset as a whole. Fine details deviating significantly from the mean object shape may not be represented by the first $c << N_T$ eigenshapes. 

Regarding the average time per image needed until convergence of the model coefficients to the local optimum, the deep shape model needed ca. 20 seconds, or 29\% of eigenshape’s 70 seconds at 32 shape coefficients, and 34 seconds, or 36\% of eigenshape’s 96 seconds at 64 shape coefficients. At the most complex model level of 128 shape coefficients, the optimization times were more equalized at ca. 70 seconds for dcgan, representing 71\% of eigenshape’s 98 seconds (see Fig.~\ref{fig:dcgan_eig_comparison_bottomRow_part1}). The increase in computation time is due to both the model complexity (i.e.~more variables), and a higher number of iterations needed to reach convergence. Both shape evolution methods are 1-2 orders of magnitude slower than the baseline Mask R-CNN solution, due to the need to solve the complex, multi-variable optimization problem in an 'on-line' fashion.

\subsubsection{\emph{Experiment 3:} Entire pipeline - baseline comparison}

\begin{figure*}[ht]
  \centering
  \includegraphics[width=0.99\linewidth]{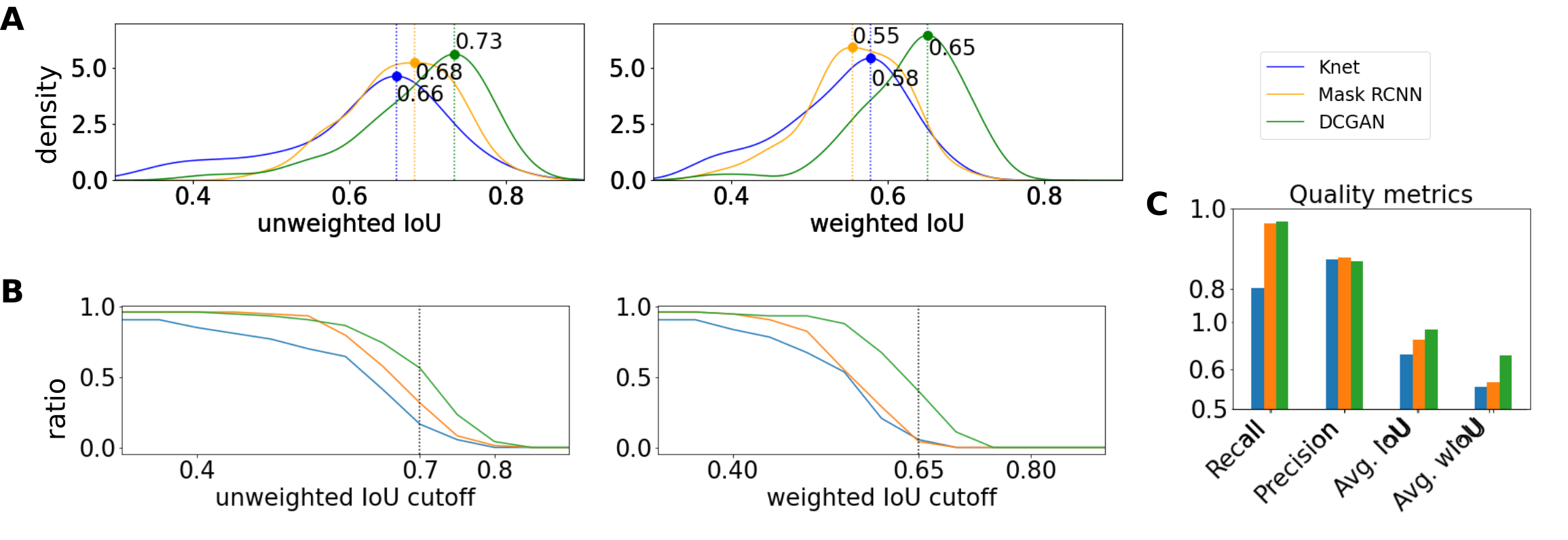}
        \caption{Quantitative comparison for various metrics on test data between 3 methods: 
 the baselines \textcolor{orange}{Mask RCNN}, \textcolor{blue}{K-Net}, and the \textcolor{DarkGreen}{DCGAN}-based version of our method. (A) Comparison of weighted and unweighted IoU distributions. Our method achieves an advantage of 2.7 pp in unweighted IoU over the best baseline, but the gain in weighted IoU is 7.8 pp, confirming that most of the improvements are localized near the contours. (B) ratio of test polygons with an
un-/weighted IoU of x or more (IoU cutoff is x axis). DCGAN reconstructs polygons with high weighted IoUs ($\geq 0.65)$ much more frequently than baselines, e.g.~ 40 \% of our method's reconstructed polygons have a weighted IoU value of 0.65 or more, while the best baseline achieves a 5 \% rate (see black dashed line on (B) right). (C) object-level precision is similar for all methods, but K-Net trails the other method by about 15 percentage points in recall, indicating that it sometimes completely misses entire tree crowns (see Fig.~\ref{fig:sampleSegmentedResults_3methods} for an example). }
 \label{fig:resultsCmpBaselines}
\end{figure*}

\begin{figure}[ht!]
  \centering
  \includegraphics[width=0.9\linewidth]{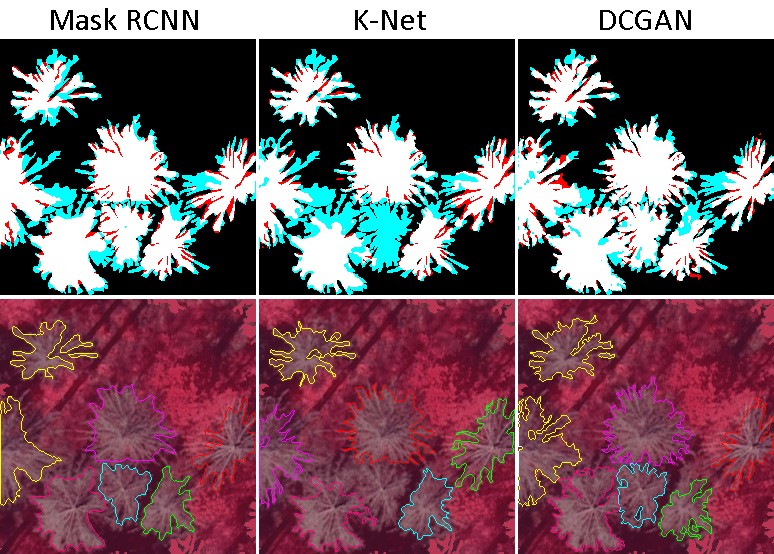}
  \caption{Visual comparison of segmented contour quality for two baseline approaches: Mask RCNN (left), K-Net (center) vs.~our method (DCGAN, right). Top: comparison of ground-truth segmented result masks: \colorbox{black}{\color{white}white}, \colorbox{black}{\color{cyan}cyan}, \colorbox{black}{\color{red}red} pixels denote respectively agreement, omission error (ground-truth only pixel), commission error (model-only pixel). Bottom: segmented contours over CIR image (random outline colors). Mask RCNN produces simpler contours, but manages to recover each tree crown. K-Net produces more complex and accurate contours, but often misses entire objects or big swaths. Our approach inherits good initial object positions from Mask RCNN and produces best overall contours.}
  \label{fig:sampleSegmentedResults_3methods}
  \end{figure}

To enable a fair and unbiased comparison of our pipeline with the K-net and Mask RCNN baselines, we used only manually labeled polygons in a 2:1 train/test split as per Section~\ref{sec:groundTruth}. Each of the baselines was trained on the same portion of the manual data, and was given ample GPU time to converge. For the sake of fairness, the training data was augmented by rotating by multiples of 15 degrees to match the input data for the Deep Shape Model. The positions obtained from the Mask RCNN run were used as the input object locations to the Deep Shape Model-based contour evolution. The binary object masks returned by each method were saved and evaluated as per the pixel-level and object-level metrics. The results are summarized in Fig.~\ref{fig:resultsCmpBaselines}. Observing the IoU and contour distance-weighted IoU distributions (Fig.~\ref{fig:resultsCmpBaselines} (a)), we note a significant shift in the per-method curves: our contour evolution attains a gain of 4 and 8 percentage points over the baselines respectively for the median unweighted and weighted IoU. Once again, the biggest gain is achieved near the contour. The baseline contour complexity shows significant differences per method. As expected, Mask R-CNN produces the coarsest contours, due to known limitations such as the loss of fine contour details due to pooling operation as well as relying on coarser feature maps from multiple layers of the feature pyramid~\citep{9577876}. On the other hand, K-Net was able to produce partial contours with a higher level of detail, however it often missed entire objects or large parts of tree crowns (see Fig.~\ref{fig:sampleSegmentedResults_3methods}). The lack of large swaths of target class image pixels in the output suggest that K-net was not able to learn the low-level task of distinguishing between dead tree pixels and the background as successfully as e.g.~Mask R CNN.

\begin{figure}[h!]
\centering
		\includegraphics[width=0.9\columnwidth]{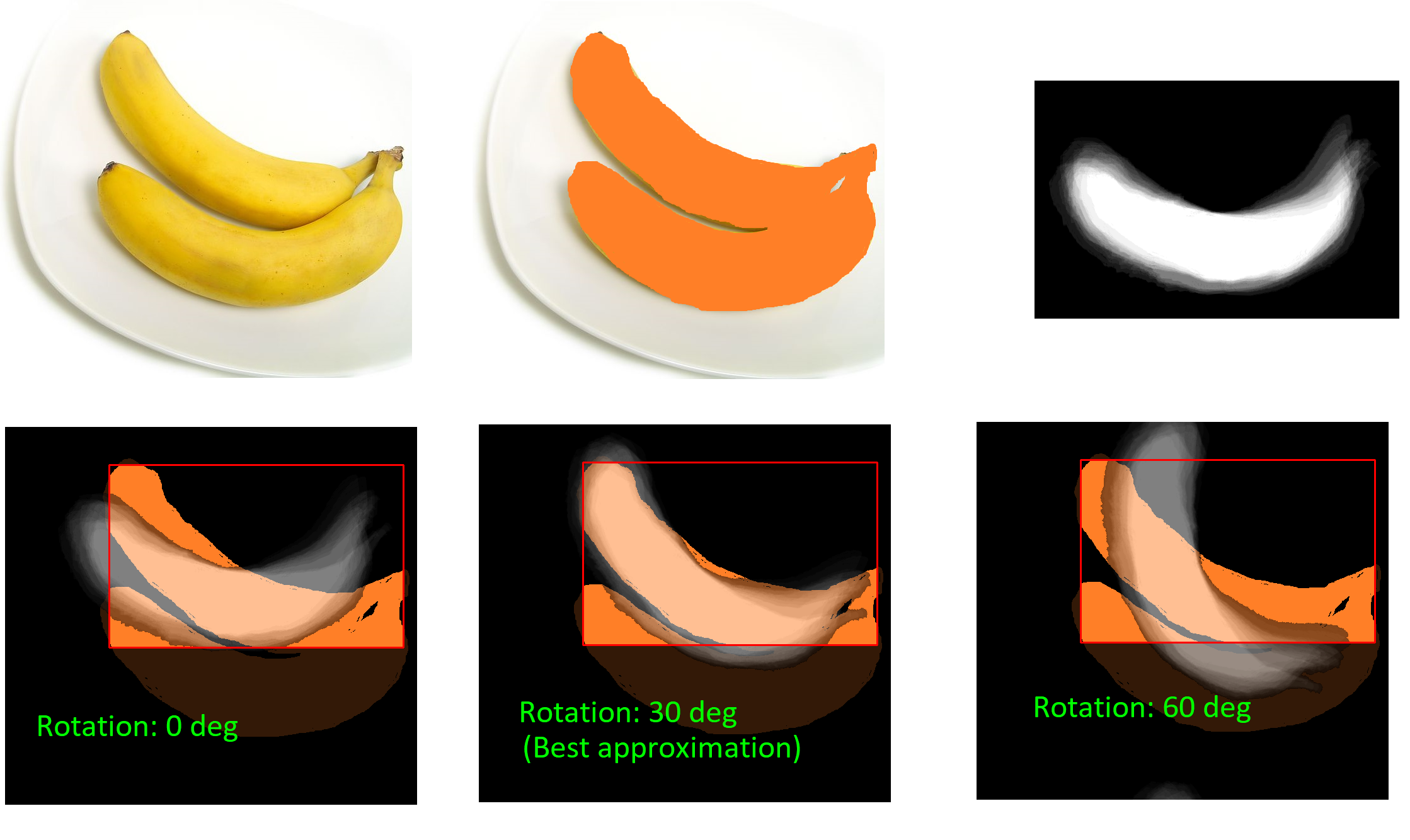}
	\caption{Searching for the optimal initial rotation parameter. Top - left: input image, center: semantic segmentation (class 'banana' shown in orange), right: initial object mask of class 'banana' in standard position. Bottom - intersection of initial mask rotated by 0, 30, and 60 degrees with the mask from semantic segmentation constrained by bounding box from object detection (red rectangle). The rotation of 30 degrees shows the best alignment of the rotated mask and the semantic segmentation map, therefore $\kappa=30^\circ$ is taken as the initial rotation parameter.}
	\label{fig:initRotationSearch}
\end{figure}

\section{Discussion and Conclusions}

    In this work, we proposed an instance segmentation method for natural resource monitoring applications in aerial imagery. 
    We consider the setting wherein the scale of the object is known and the problem lies in delineating precise contours of multiple adjacent instances belonging to few categories. 
    Whereas current segmentation methods are very successful with broad-spectrum class \textit{recognition} on \textit{terrestrial-view} high-resolution images, (e.g. benchmark datasets such as COCO, ImageNet), they face well-known limitations in delineating detailed contours, e.g. Mask RCNN can only consider down-sampled low-resolution images.
    To overcome this deficit, we leverage the power of \textit{active shape models}, namely level-set methods, for the \textit{evolution} of precise individual contours while simultaneously exploiting the versatility of neural network architectures.
    
    
    %
    %
    
    We formulate the problem as Bayesian maximum a posteriori inference which incorporates shape, location, and position priors from state-of-the-art CNN architectures within a \textit{contour evolution framework}. Furthermore, central to this work is the extension of the classic Eigenshape active contour model to the non-linear setting. This is accomplished in two ways: (1) it allows for arbitrary nonlinear kernels to be applied to the shape data, possibly defining a different topology of the shape coefficient space, and (2) it extends the Eigenshape's linear representation of the signed distance function in terms of a linear combination of eigenmodes, and instead uses a convolutional network to \emph{decode} the shape coefficients into a shape mask, much like a GAN 'decodes' latent variables into a generated image. This avoids the potentially intractable problem of finding eigenvectors in the possibly infinite-dimensional reproducing kernel Hilbert space associated with the chosen non-linear kernel. Our modifications open up possibilities of finding tailored kernels as well as decoder network architectures for specific problems.



    The Bayesian framework is only loosely coupled with specific neural network architectures that generate prior information for the shape, location, and appearance modalities. This allows for simple drop-in replacement of new networks and shape models as better architectures become available. Moreover, our framework has an end-to-end GPU implementation, which enables the use of modern computational hardware and scalability.

    In extensive numerical experiments, we compare our method with Mask R-CNN and K-net for delineation of precise tree crowns in color infrared aerial imagery of dead tree crowns within the Bavarian Forest National Park. 
    Our primary findings are: 
    (1) our framework significantly improves the quality of the reconstructed object boundaries compared to two state-of-the-art general purpose segmentation networks on a set of challenging forest scenes with multiple adjacent and overlapping tree crowns, and 
    (2) the deep shape model extension (shape prior) outperforms the Eigenshape baseline both in terms of the reconstructed contours and execution time.

It is worth noting that, unlike foundation-model-based segmentation approaches that typically require prompt engineering or large-scale pretraining data to generalize~\citep{Xu2026,QUE2026111162,NEURIPS2025_0cb66bdf}, our framework explicitly models shape, location, and appearance priors within an interpretable Bayesian energy formulation, and can be trained with substantially fewer annotated examples (200 manually labeled polygons in this study). This makes it particularly suitable for specialized ecological monitoring tasks where large annotated datasets are unavailable — a common constraint in precision forestry applications distinct from general-purpose benchmarks such as COCO. Note that a direct experimental comparison with recent SAM-based tree crown segmentation approaches was not included in this study, as these methods rely on prompt-based or promptable-mask interaction paradigms that are not directly compatible with the fully automatic, training-set-driven evaluation protocol adopted here. Nevertheless, since our Bayesian framework treats the semantic segmentation and object detection modules as interchangeable black-box priors, replacing them with SAM-derived or foundation-model-derived probability maps constitutes a natural and technically straightforward extension.

    In future work, we will translate optimizations from the classic level set/contour evolution domain to the GPU setting, e.g.~focusing only on a restricted pixel band around each evolving contour, which will greatly reduce computational demands. 
    Furthermore, in place of the linear kernel function considered here, we will explore more sophisticated, non-linear kernel functions in the shape model. Beyond these algorithmic extensions, recent large-scale benchmark efforts — including a nationwide dataset for individual tree mortality monitoring~\citep{NEURIPS2025_f2262528} - offer promising opportunities for validating our framework at broader geographic and ecological scale. Furthermore, given the rapid emergence of vision foundation models tailored to forestry applications~\citep{Xu2026,bountos2025fomo}, a direction for future work is to substitute the semantic-segmentation and object-detection priors with representations derived from such pretrained foundation models, while retaining the loosely-coupled Bayesian energy formulation that is central to our approach.


\section*{Declaration of competing interest}
The authors declare that they have no known competing financial interests or personal relationships that could have appeared to influence the work reported in this paper.

\section*{Acknowledgements}
We would like to thank Dr. Konrad Schindler for his valuable input, which greatly strengthened this work. This work was also supported by National Natural Science Foundation of China (Grant No.42171361)

\appendix
\section{Initialization of shape orientation parameter}\label{sec:appendixInitParam}

For instance segmentation settings which include an active and meaningful orientation term, a somewhat more elaborate initialization procedure is needed. An estimate of the initial object orientations $\tilde{\kappa}_i$ is now required alongside the shape coefficient vectors. The initialization of the latter must also consider the orientation prior $\mathcal{P}_{rot}$, and the fact that the shape model is not rotation invariant, but rather has been trained on shapes in a 'standard' orientation. More formally, we consider the one-variable optimization problem for each shape $i$ separately:
\begin{equation}
\label{eq:initRotation}
\begin{split}
M^{rot}_i(\kappa) &\equiv R(\kappa) \cdot M^{init}_i\\
M^{dec}_i(\kappa) &\equiv H(f_d(\Pi^{\Phi}_{V^1}(M^{rot}_i(\kappa),\ldots,\Pi^{\Phi}_{V^c}(M^{rot}_i(\kappa))))\\
\max_{\tilde{\kappa}_i} E^{rot}
_i(\tilde{\kappa}_i) &= \log \mathcal{P}_{rot}(\tilde{\kappa}_i|c_i) - \mathcal{H}(M^{dec}_i(\tilde{\kappa}_i), M^{rot}_i(\tilde{\kappa}_i))
\end{split}
\end{equation}

The above equation, illustrated by Fig.~\ref{fig:initRotationSearch}, represents the search for the optimal initial orientation value $\tilde{\kappa}_i$, which maximizes the sum of the orientation log-prior and the negated reconstruction error resulting from decoding the projection of the init mask $M^{init}_i$ rotated by kappa onto the shape model's principal components back into image space using the trained decoder function $f_d$. Specifically, $M^{rot}_i(\kappa)$ denotes the init mask $M^{init}_i$ transformed using a rotation matrix $R(\kappa)$ by angle $\kappa$. Also, $M^{dec}_i$ denotes the \emph{decoded} projection $(\Pi^{\Phi}_{V^1}(M^{rot}_i(\kappa), \ldots)$ of $M^{rot}_i(\kappa)$ onto the shape model's principal components $V^1,\ldots,V_c$. The decoded image, obtained by applying the decoder $f_d$ (see Sec.~\ref{sec:generalizing_eigenshape_rep}), is then binarized using the Heaviside function $H$ to enable a comparison with the rotated mask $M^{rot}_i$. Finally, the term $\mathcal{H}(M^{dec}_i(\tilde{\kappa}_i), M^{rot}_i(\tilde{\kappa}_i))$ refers to the cross-entropy between $M^{dec}_i$ and $M^{rot}_i$, and can be understood as measuring the reconstruction error of the round-trip encoding-decoding of the rotated mask $M^{rot}_i(\tilde{\kappa}_i)$. To gain some insight into the computational cost of solving the problem given by Eq.~\ref{eq:initRotation}, we will discuss the computational complexity of its components, considering the following quantities as variables that influence the complexity: (i) the number of image elements (pixels) in the binary mask, $N_{pix}=D_{c_i}^2$, (ii) the number $c$ of top principal components in the shape model, (iii), the number $N_T$ of training shapes used for learning the shape model and possibly the orientation prior $\mathcal{P}_{rot}$. Evaluating $E^{rot}_i$ for a particular value of $\tilde{\kappa}_i$ can be broken down in the following chain of operations:
\begin{enumerate}
   \item Retrieving the value of the orientation prior $\mathcal{P}_{rot}$ at $\tilde{\kappa}_i$ - complexity is $O(1)$ for a parametric model, or $O(N_T)$ for a non-parametric model, e.g.~\emph{KDE}.
   \item Rotating the initial mask $M^{init}_i$ by angle $\tilde{\kappa}_i$ - complexity is $O(N_{pix})$
   \item Projecting the rotated mask onto the top $c$ eigenvectors of the shape model. This involves evaluating the shape kernel $\mathcal{K_T}$ $O(N_T)$ times, each of which involves $O(N_{pix})$ operations
   \item Decoding the projected coefficients with the feed-forward network $f_d$. The computational complexity depends on the network architecture. For example, our proposed decoder architecture based on fully convolutional networks scales linearly with the image size  $O(N_{pix})$ - see Sec.~\ref{sec:deepShapeModels}
   \item applying the Heaviside function $H$ and the cross-entropy $\mathcal{H}$ both incur a computational cost of $O(N_{pix})$ 
\end{enumerate}

The total cost of evaluating $E^{rot}_i$ is dominated by the cost of computing the kernel function for all training examples $N_T$. Depending on the training data size, it might be prohibitively expensive. This could be alleviated by clustering of training data in the low-dimensional shape coefficient space and selecting a representative subset of the training data to construct a more compact shape model.

In order to ensure that the initialization procedure remains simple and light-weight, we propose to perform a line search to minimize $E^{rot}_i$ on the range of valid orientation angles (e.g.$[0;2\pi])$, with an equal-interval step size of $\Delta\kappa$. There is no need to solve Eq.~\ref{eq:initRotation} to a greater precision, since the initial values are only the starting points of the true optimization which occurs simultaneously over all shapes.

\section{Smooth maximum approximation}\label{appendix:smoothMax}

The discrete maximum operator over $(x_1,\ldots,x_D)$ can be approximated by a function $\tilde{S}_{\gamma}(x_1,\ldots,x_D)$ parameterized by the value $\gamma$, such that $\tilde{S}_{\gamma}(x_1,\ldots,x_D) \to \max (x_1,\ldots,x_D)$ as $\gamma$ approaches infinity. There are several choices for $\tilde{S}_{\gamma}$ in literature, including:
\begin{itemize}
    \item $\tilde{S}^{avg}_{\gamma}(x_1,\ldots,x_D)=\frac{\sum_i x_i\exp(\gamma x_i)}{\sum_i \exp(\gamma x_i)}$, the exp-weighted average
    \item $\tilde{S}^{lse}_{\gamma}(x_1,\ldots,x_D)=\frac{1}{\gamma}\log[\exp(\gamma x_1) + \ldots + \exp(\gamma x_D)]$, the log-sum-exp function
    \item $\tilde{S}^{pn}_{\gamma}(x_1,\ldots,x_D)=(\sum_i |x_i|^\gamma)^\frac{1}{\gamma}$, the $p-$norm
\end{itemize}
The quality of the approximation improves with increasing $\gamma$. On the other hand, in practice the value of $\gamma$ is limited by the precision of floating point number implementations in hardware due to potential issues of numerical stability when performing the logarithm and exponentiation operations.

\section{Foreground membership function - multiple object classes}

The single-foreground class formulation introduced in the previous section takes advantage of the fact that in a binary setting, the 'probabilities' of a pixel's assignments to the foreground and background classes sum to one, and therefore both may easily be expressed using the approximate maximum 'membership' of the pixel in the foreground class, i.e.~$\tilde{S}[\tilde{H}(\phi_1)[\omega],\ldots,\tilde{H}(\phi_{N_o})[\omega]]$. However, in the multi-class case this no longer holds, and a background class membership needs to be explicitly defined. As there are multiple ways of defining the background probability based on the foreground classes, we base our choice on the intuition that the background probability should be high if \emph{all} level-set functions of foreground objects are low, and low if \emph{any} foreground class membership is high. We adopt the following notation for the per-class maximum of level-set functions:
\begin{equation}
    S^{i} \equiv \max_{k: c_k = i} \tilde{H}(\phi_k), i=1\ldots |\mathcal{C}|
\end{equation}
Thus, the background class 'membership' function $S^0$ meeting our requirements can be defined as:

\begin{equation}
\begin{split}
    S^{0} &\equiv \min_{k=1}^{|\mathcal{C}|} (1 - S^k)=1-\max _{k=1}^{|\mathcal{C}|} S^k\\
    &=1-\max_{k=1}^{|\mathcal{C}|} (\max_{l: c_l = k} \tilde{H}(\phi_l))\\
    &=1-\max_{k=1}^{N_o} \tilde{H}(\phi_k)
\end{split}
\end{equation}

Replacing the $\max$ operator with its smooth approximation $\tilde{S}$, we can form pseudoprobabilities of class assignments at any image element $\omega$ as $\tilde{p}^i[\omega]=\tilde{S}^{i}[\omega] / \sum_{k=0}^{|\mathcal{C}|}\tilde{S}^{k}[\omega]$. This gives rise to a natural formulation of the multi-class cross-entropy:
\begin{equation}
\begin{split}
    &E_{img} = -\int_{\omega} \frac{1}{\sum_{k=0}^{|\mathcal{C}|}\tilde{S}^{k}[\omega]} \sum_{i=0}^{|\mathcal{C}|} \tilde{S}^{i}[\omega] \log \mathcal{P}_{\text{sem}}(c_i|\omega)d\omega
\end{split}
\end{equation}

Note that owing to the use of the smooth approximations $\tilde{H}, \tilde{S}$, the image energy $E_{img}$ is differentiable with respect to the level-set functions $\phi_i$, and, via the chain rule, to the parameters $\pi_i$ of the evolving shapes.

\bibliographystyle{elsarticle-harv} 
\bibliography{ref}





\end{CJK}
\end{document}